%% file: main.tex
\documentclass[journal]{IEEEtran}

\usepackage{hyperref, url, booktabs, amsfonts, nicefrac, microtype, amsmath, amssymb, graphicx, multirow, mathtools, caption, subcaption, lipsum, multicol, color,cite}       

\begin{document}
\title{Harvesting Ambient RF for Presence Detection Through Deep Learning}

\author{%
	Yang~Liu, Tiexing~Wang, Yuexin~Jiang and Biao~Chen,~\IEEEmembership{Fellow,~IEEE}
	\thanks{Y. Liu, T. Wang and B. Chen are with the department of Electrical Engineering and Computer Science, Syracuse University, Syracuse, NY, 13244, USA. Y. Jiang was with the department of Electrical Engineering and Computer Science, Syracuse University, Syracuse, NY, 13244. He is now with IFR Technologies LLC, Syracuse, NY, 13244. The work was supported in part by a gift fund from Syracuse Research Corporation, Syracuse, New York and by the National Science Foundation under grant CNS-1730882.

Source code and WiFi data repository can be found on github at \url{https://github.com/bigtreeyanger/presence_detection_cnn}.}
}
\maketitle

\begin{abstract}
\input{modules/abstract.tex}
\end{abstract}
\begin{IEEEkeywords}
Presence detection, WiFi sensing, MIMO-OFDM, convolutional neural networks, channel state information.
\end{IEEEkeywords}

\section{Introduction}
\input{modules/introduction}

\section{MIMO-OFDM System model}\label{sec::model}
\input{modules/system_model}

\section{System design}\label{sec::design}
\input{modules/system_design}

\section{Experiment Setup}\label{sec::experiment}
\input{modules/experiment}

\section{Conclusion}\label{sec::conclusion}
\input{modules/conclusion}




\bibliography{reference}
\bibliographystyle{IEEEtran}
\end{document}

%% file: modules/abstract.tex
This paper explores the use of ambient radio frequency (RF) signals for human presence detection through deep learning. Using WiFi signal as an example, we demonstrate that the channel state information (CSI) obtained at the receiver contains rich information about the propagation environment. Through judicious pre-processing of the estimated CSI followed by deep learning, reliable presence detection can be achieved. Several challenges in passive RF sensing are addressed. With presence detection, how to collect training data with human presence can have a significant impact on the performance. This is in contrast to activity detection when a specific motion pattern is of interest. A second challenge is that RF signals are complex-valued. Handling complex-valued input in deep learning requires careful data representation and network architecture design. Finally, human presence affects CSI variation along multiple dimensions; such variation, however, is often masked by system impediments such as timing or frequency offset. Addressing these challenges, the proposed learning system uses pre-processing to preserve human motion induced channel variation while insulating against other impairments. A convolutional neural network (CNN) properly trained with both magnitude and phase information is then designed to achieve reliable presence detection. Extensive experiments are conducted. Using off-the-shelf WiFi devices, the proposed deep learning based RF sensing achieves near perfect presence detection during multiple extended periods of test and exhibits superior performance compared with leading edge passive infrared sensors. Comparison with existing RF based human presence detection also demonstrates its robustness in performance, especially when deployed in a completely new environment. The learning based passive RF sensing thus provides a viable and promising alternative for presence or occupancy detection.

%% file: modules/introduction.tex
Presence detection plays a key role in improving operation efficiency and reducing carbon footprint for office and residential buildings. The use of occupancy information in controlling 
HVAC and lighting systems has become increasingly prevalent. 
Existing methods for human presence detection include passive infrared (PIR), microwave, CO$_2$, wearable sensors, and cameras \cite{teixeira2010survey}, among others. Microwave sensors are overly sensitive as they tend to have frequent false alarms, e.g., detecting movements outside of intended coverage areas. CO$_{2}$ sensors have a slow response time and a high cost barrier. Cameras raise privacy concerns and are sensitive to lighting conditions. Wearable sensors/devices can be intrusive or cumbersome for users. PIR sensors are the most widely deployed method for presence detection. PIR sensors pick up infrared emission using its on-board pyroelectric sensor and detect movement of humans (or objects) through heat variation within the field of view. Its drawback is its low sensitivity and limited coverage and it is mostly used for isolated lighting control.

This paper explores the use of radio frequency (RF) signals for presence detection. In particular, we use WiFi signals in the current work given its ubiquity in almost all indoor environment. Current and future WiFi systems (i.e., the upcoming WiFi-6) employ multiple-input and multiple-output orthogonal frequency-division multiplexing (MIMO-OFDM) at the physical layer. As such, the channel state information (CSI) contains rich information about the ambient environment in spatial, temporal, and frequency domains.

Exploiting ambient RF signals for situational awareness tasks such as detecting, localizing, tracking, and identifying human motion/activities have been extensively studied in the literature~\cite{abdelnasser2015wigest, gu2015paws, moussa2009smart, palipana2018falldefi, xu2019indoor, zheng2017design, li2017ar, liu2014wi,wang2014eyes,arshad2017wi, wang2017device,gao2017csi,xu2018time, li2017indotrack, qian2017widar,soltanaghaei2017peripheral, zhao2013radio,wu2015deman, ma2018signfi,zou2017freedetector, qian2018pads, gong2016adaptive, palipana2016nonlinear, xiao2012fimd, zhu2017r, kianoush2019people, savazzi2019use, holl2017holography}. 
Early work for indoor RF sensing mainly relies on 
received signal strength indicator (RSSI)~\cite{abdelnasser2015wigest, gu2015paws, moussa2009smart}.
RSSI measures instantaneous attenuation of RF signals at the receiver and its temporal variation can be associated with motion/activities of humans/objects. 
Recently, more fine-grained features such as CSI 
have been used for RF sensing. 
For example, different human activities (e.g., running, walking and eating) or locations can be recognized by analyzing their unique effect on the CSI~\cite{wang2014eyes, arshad2017wi, wang2017device,gao2017csi,xu2018time, li2017indotrack, qian2017widar}. 
Other interesting applications, e.g., sign language classification~\cite{ma2018signfi} and human identification~\cite{wang2016gait, zou2018wifi, regani2019driver}, have demonstrated that the CSI from WiFi signals contains subtle and important features in the propagation environment.

There is an important distinction between presence detection and activity detection (e.g., sign language~\cite{ma2018signfi} or fall detection~\cite{Wang2017Fall, palipana2018falldefi}). For detection of particular activities, one can use either a pattern-based or model-based approach~\cite{wu2017device, wang2019survey, guo2017behavior} as a target activity often imposes identifiable signatures on RF propagation. Thus, hand-crafted features extracted from received signals can be exploited~\cite{xu2019indoor, wang2014eyes} or one can try to establish a CSI model for the target activity~\cite{xin2018freesense, li2017indotrack, wang2015csi-speed}. 
Alternatively, a data-driven approach can be used where collected training data are fed to machine learning algorithms (e.g., a neural network) for discriminating different states (labels) through training. For presence detection, however, there is no defined activities when humans are present thus a pattern-based or model-based approach is typically not adequate. While a data-driven approach appears to be a natural choice, it is unclear {\em a priori} what would be the best way to collect training data for presence detection. Perhaps the only reasonable assumption is that humans are not expected to be completely still for an extended period of time. While there exist studies on presence detection using RF signals through either carefully calibrating the human free environment ~\cite{soltanaghaei2017peripheral, zhao2013radio} or breathing detection~\cite{wu2015deman, zeng2019farsense}, their performance is highly sensitive to environment change (e.g., room change, furniture move) and human locations (as in the case of breathing detection).


There is prior work on presence detection using CSI. In pattern-based methods, hand-crafted features have been extracted to capture distinctive signal patterns introduced by human presence. Among them, the correlation matrix of CSI time series is most widely explored~\cite{qian2018pads, zhu2017r, xiao2012fimd, gong2016adaptive, palipana2016nonlinear}. Features such as eigenvalues/eigenvectors are used as an input to a classifier such as a support vector machine (SVM) to find decision boundaries. Other suggested features include periodicity after performing continuous wavelet transform~\cite{lv2019robust}, temporal similarity of CSIs across frequencies~\cite{zou2017freedetector}, histograms of CSI amplitude~\cite{zhou2013omnidirectional}, and statistics from average Doppler spectrum~\cite{di2018doppler}. These pattern-based approaches are susceptible to inconsistent human activities, i.e., when mismatch occurs between human motion and those stored in the profile~\cite{wu2017device} and are thus prone to performance degradation and limited coverage~\cite{qian2018pads, zhou2013omnidirectional}.  
Model-based methods \cite{li2018training-free, xin2018freesense, li2019real} often exploit estimated path change to represent Doppler effect induced by moving humans.
The performance is highly sensitive to conditions such as human movement direction, number of people in the monitored region, and other ambient factors~\cite{li2019real,xin2018freesense}. 

Alternatively, data-driven approaches employing learning systems (e.g., neural networks) have also been proposed. Examples include feeding CSI amplitude and phase to multiple layer perceptions (MLP) to classify three common human modes - absence, working and sleeping~\cite{fang2019enhanced}, and Person-in-WiFi~\cite{wang2019person} that tries to find mapping from CSIs to 2D human image. 
However, an acute challenge in using deep neural networks is overfitting due to the large amount of trainable parameters compared with often limited training data. This problem becomes even more severe in RF sensing as the wireless environment can evolve over time or change dramatically for different rooms, leading to significant performance degradation. As reported in~\cite{fang2019enhanced}, deploying the pre-trained model to a new environment has resulted in a drop of accuracy from $94.5\%$ to around $50\%$. Similar performance losses have been reported in other systems, see, e.g., \cite{wang2019person}. 


This paper proposes a WiFi CSI based presence detection system consisting of pre-processing for data representation, a convolutional neural network (CNN) for motion detection, and post-processing for the eventual presence detection. The rationale of choosing CNN is its ability to exploit CSI variation in multiple dimensions thanks to the MIMO-OFDM waveforms at the physical layer. 
In contrast, while recurrent neural network (RNN) can also be used for learning variation pattern of CSI over time, it introduces more computational overhead compared with CNN and the long-term memory offered by RNN provides no meaningful gain in this application given that channel correlation in time and frequency both diminishes as distances increase. With MLP, on the other hand, local features, i.e., change in correlation in both temporal and frequency domains, are not explicitly explored hence MLP typically requires much deeper networks to achieve similar performance.

\subsection{Summary of Contributions}
The proposed parallel CNN architecture separates CSI magnitude and phase in the learning system. 
This is much more robust than stacking up real and imaginary components as it allows for different pre-processing of CSI magnitude and phase. This is crucial in exploiting motion induced CSI variation in the presence of various channel and hardware impairments. In particular, it allows us to capture motion induced information in CSI phase as opposed to discarding the phase information due to inherent impediments such as carrier frequency offset (CFO) and sampling time offset (STO)~\cite{lv2017robust, zhu2017r, zhou2013omnidirectional, gong2016adaptive, xiao2012fimd, palipana2016nonlinear}.  

Pre-processing of CSI estimate is carefully designed where spatial, temporal, and frequency domain information is exploited in a holistic manner. The pre-processing takes into consideration how human movement affects CSI while insulating against unintended distortion in RF circuitry (e.g., CFO/STO). Fourier transform is used to localize important motion-induced features in the constructed image, making it more amenable for presence detection as CNN builds its ability for discriminating data through local features (i.e., small kernel size). 

An important contribution of the present work is extensive test implemented using commercial off-the-shelf (COTS) WiFi devices. Assuming that humans are not completely still for an extended period of time, our presence detection compares much more favorably against that of commercial PIR sensors. We note that the comparison is done without customized WiFi hardware. For example, we have found that when using USRP systems~\cite{usrp}, which we have full control on sampling frequency and gives a much cleaner CSI estimate, performance can improve substantially over COTS WiFi receiver.

\subsection{Organization and Notation}
Section~\ref{sec::model} describes the MIMO-OFDM waveforms and how human movement impacts wireless channels. The design of the sensing system including  pre-processing, the proposed CNN architecture, and post-processing is described in Section~\ref{sec::design}. Experiment setup and the corresponding test results are provided in Section \ref{sec::experiment} followed by conclusion in Section~\ref{sec::conclusion}.

Scalars are denoted by either lower or upper case letters, e.g., $a$ and $A$. Column vectors and matrices are denoted by lower and upper case bold letters, e.g., $\mathbf{a}$ and $\mathbf{A}$. The $i$-th entry of a vector $\mathbf{a}$, the $(i,j)$-th entry of $\mathbf{A}$, and the $(i,j,k)$th entry of a $3$-D array $\mathbf{A}$ are denoted by $\mathbf{a}_{i}$, $\mathbf{A}_{i,j}$, and $\mathbf{A}_{i,j,k}$, respectively. 
$\mathbf{A}_{:,k}$ and $\mathbf{A}_{k,:}$ are used to denote the $k$-th column and the $k$-th row of $\mathbf{A}$. Similarly, $\mathbf{A}_{:,:,k}$ represents the $2$-D matrix at depth $k$ of the $3$-D array $\mathbf{A}$. $\mathcal{F}$ denotes the discrete Fourier transform (DFT) hence $\mathcal{F}^{-1}$ the inverse DFT. $|A|$ and $\angle A$  denote magnitude and phase of a complex number $A$.

%% file: modules/system_model.tex
\subsection{MIMO-OFDM}
Consider a MIMO-OFDM system with $N_{t}$ transmit antennas, $N_{r}$ receive antennas, and $N_{sc}$ subcarriers. Each physical layer frame consists of $M$ OFDM symbol blocks. Denote by $\mathbf{d}^{p}[m,i]$ the $m$-th frequency domain OFDM symbol vector in the $i$-th frame sent by the $p$-th transmit antenna. 
The discrete-time complex baseband signal corresponding to $\mathbf{d}^{p}[m,i]$ is given by
$\mathbf{s}^{p}[m,i] = \mathcal{F}^{-1}\left(\mathbf{d}^{p}[m,i]\right)$.
At the receiver, after cyclic prefix removal and applying DFT, the complex baseband sample at the $q$-th receive antenna in the frequency domain can be expressed as, for $k=0,1,\ldots, N_{sc}-1$, 
\begin{equation}\label{eq:channel_response}
\mathbf{y}_{k}^{q}[m,i] = \sum_{p=0}^{N_{t}-1}\mathbf{h}_{k}^{q,p}[i]\mathbf{d}_{k}^{p}[m,i]+\mathbf{v}_{k}^{q}[m,i],
\end{equation}
where $\mathbf{v}_{k}^{q}[m,i]$ is the channel noise and $\mathbf{h}_{k}^{q,p}[i]$ the channel coefficient from the $p$-th transmit antenna to the $q$-th receive antenna on the $k$-th subcarrier, assumed to be a constant within one frame (i.e., for all $M$ OFDM blocks within the $i$-th frame). Expressed in vector form, we have 
\[
\mathbf{y}_{k}[m,i] = \mathbf{H}_{k,:,:}[i]\mathbf{d}_{k}[m,i]+\mathbf{v}_{k}[m,i],
\]
where $\mathbf{H}[i]$ is the 3-D channel array of shape $N_{sc}\times N_r\times N_t$ corresponding to the $i$-th OFDM frame, $\mathbf{H}_{k,q,p}[i]=\mathbf{h}_{k}^{q,p}$, $\mathbf{d}_{k}[m,i] = \left[\mathbf{d}_{k}^{0}[m,i],\ldots,\mathbf{d}_{k}^{N_{t}-1}[m,i] \right]^{T}$ and $\mathbf{y}_{k}[m,i]=\left[\mathbf{y}_{k}^{0}[m,i],\ldots,\mathbf{y}_{k}^{N_{r}-1}[m,i] \right]^{T}$.


\subsection{Effect of Human Motion on MIMO-OFDM Channel\label{sec:motion}}

\begin{figure}[!t]
	\begin{center}
	\begin{subfigure}[b]{0.24\textwidth}
		\includegraphics[width=1\textwidth]{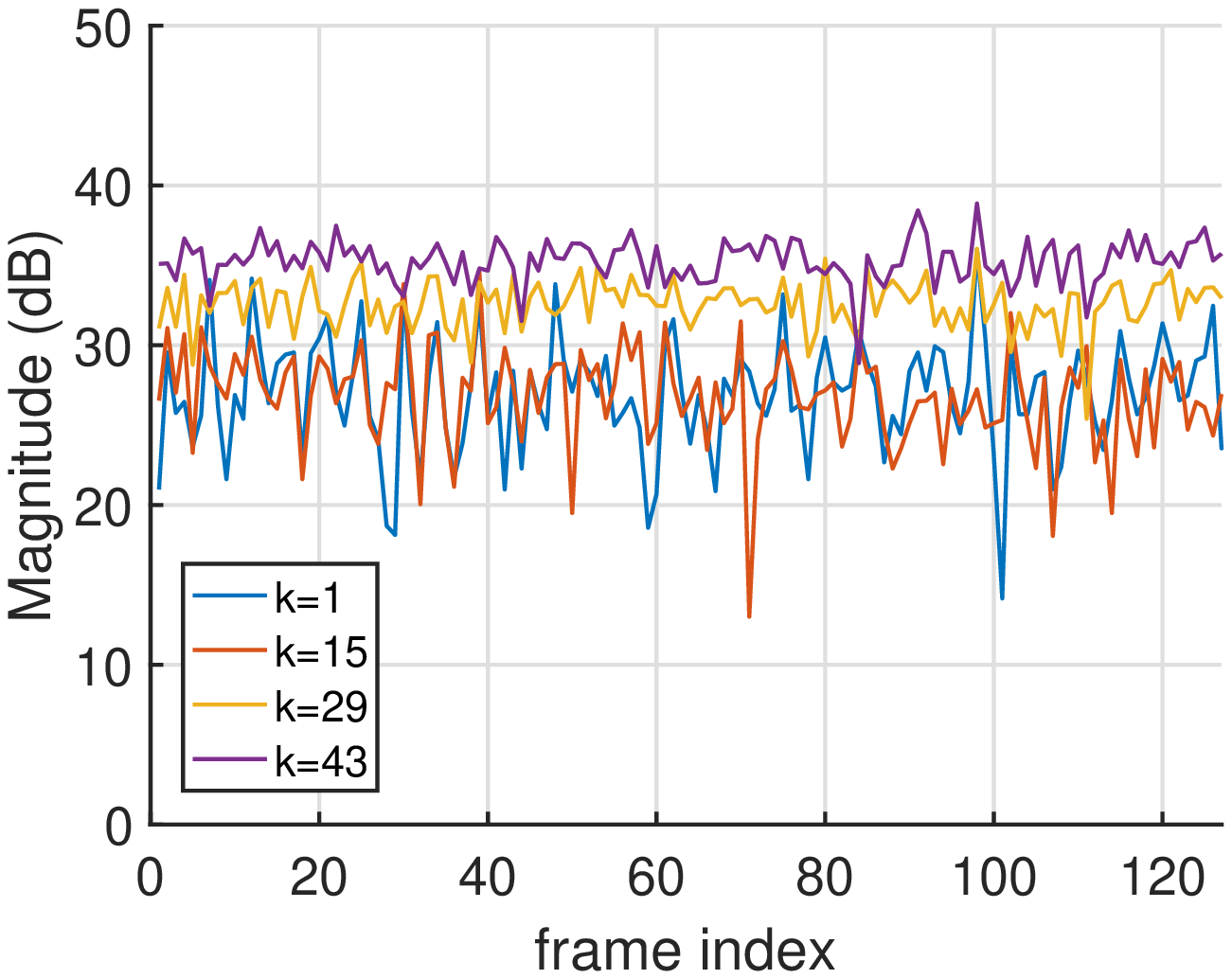}
		\caption{Human-free environment}
		\label{fig:static_env}
	\end{subfigure}	
	\begin{subfigure}[b]{0.24\textwidth}
		\includegraphics[width=1\textwidth]{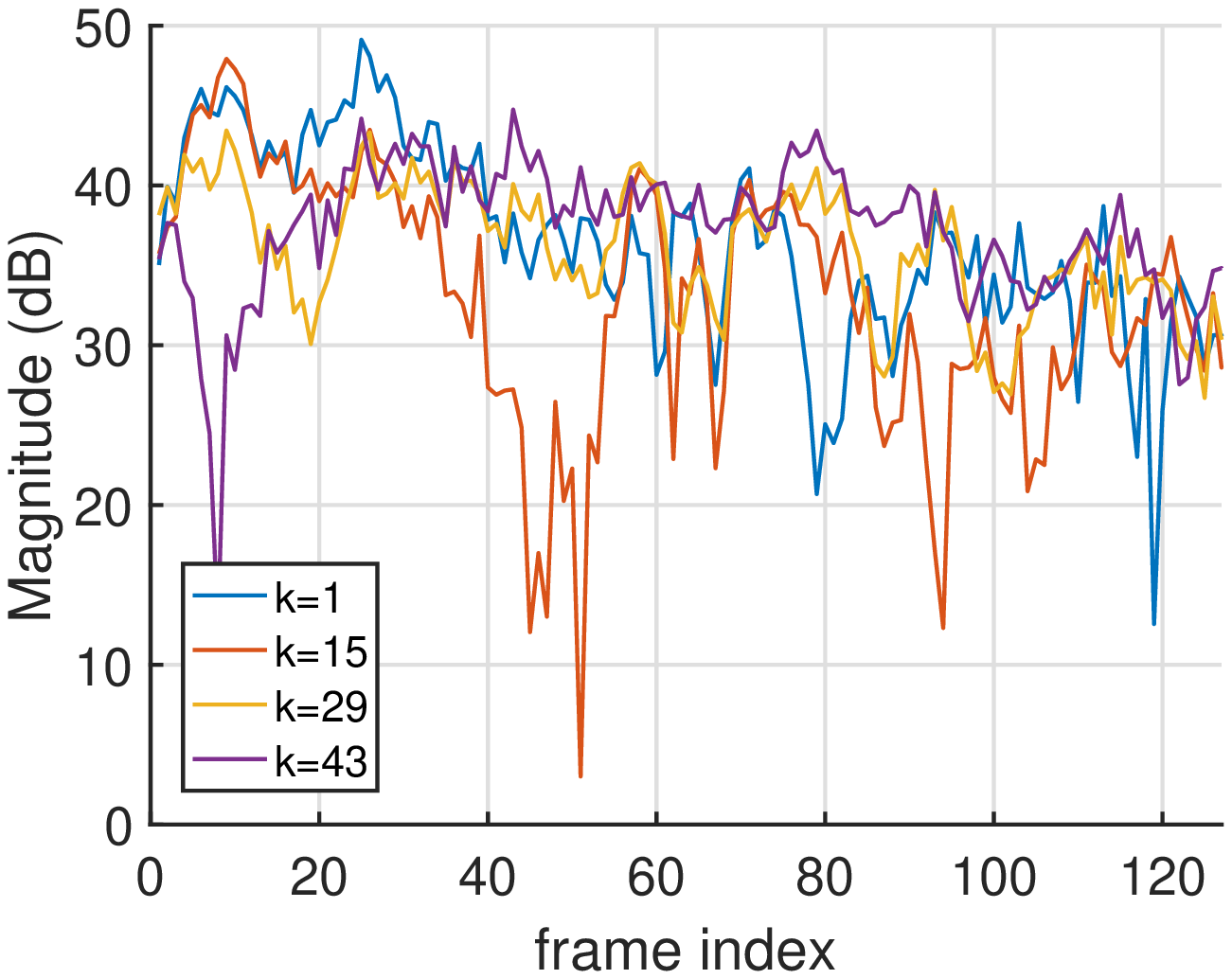}
		\caption{Human movement}
		\label{fig:motion_env}
	\end{subfigure}
	\caption{CSI magnitude variation over time for four evenly spaced subcarriers.}
	\label{fig:csi_amplitude_test}
	\end{center}
\end{figure}

Human motion leads to CSI variation in both the frequency (across subcarriers) and temporal (across frames) domains. The presence and movement of humans introduce new paths whose delays are affected by human locations, leading to change in path delay profile, hence the change in channel frequency response. Human movement also induces temporal CSI variation as signals may add in-phase or out-of-phase at the receiver depending on the locations of humans. An alternative interpretation is the increase of Doppler spread due to human movement in an otherwise static environment, leading to time-selective channel fading~\cite{Rappaport}. An example using real WiFi measurement of the variation of $\left|\mathbf{H}_{k,q,p}[i]\right|$ over frame index $i$ with and without human movement for fixed $q$ and $p$ is shown in Fig.~\ref{fig:csi_amplitude_test} for four evenly spaced subcarrriers. Clearly, with movement, channel variation both in frequency (across subcarriers) and in time (along the horizontal axis) increases. 

The effect of human movement on the CSI in the spatial dimension is more subtle. 
The human motion induced CSI variation in the temporal and frequency domains applies to every transmit-receive antenna pair. The fact that multiple transceiver pairs ({\em a.k.a}, spatial diversity) exist in the MIMO-OFDM system should be exploited for enhanced sensing performance. CNN is a natural choice for exploiting such spatial diversity by mapping temporal-frequency CSI corresponding to each transceiver pair to a layer (`channel') in a CNN architecture, much like the way colored images are processed in a CNN where RGB pixels serve as separate channels.

\begin{figure}[!t]
	\begin{center}
	\begin{subfigure}[b]{0.24\textwidth}
		\includegraphics[width=1\textwidth]{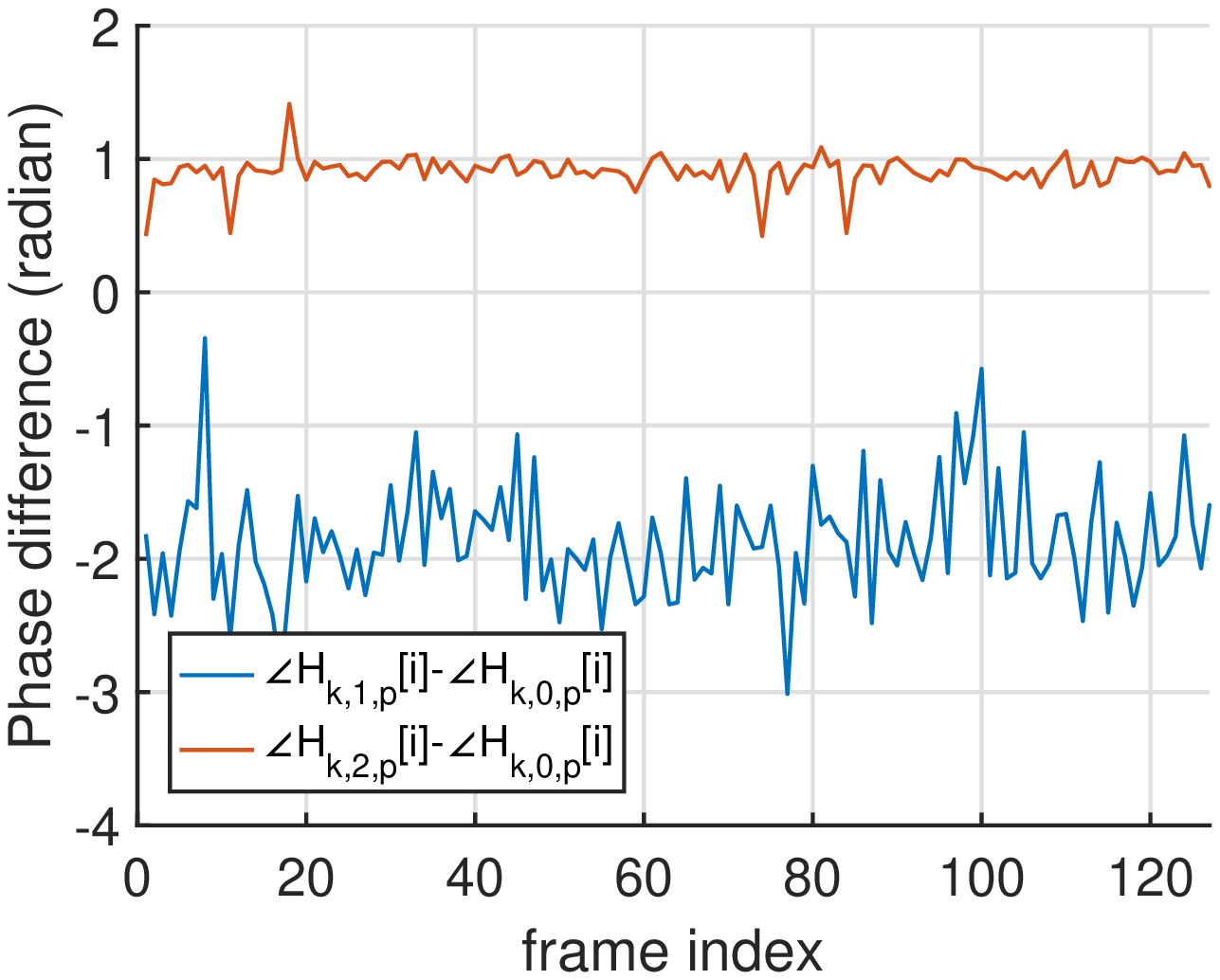}
		\caption{Human-free environment}
		\label{fig:phase_static_env}
	\end{subfigure}
	\begin{subfigure}[b]{0.24\textwidth}
		\includegraphics[width=1\textwidth]{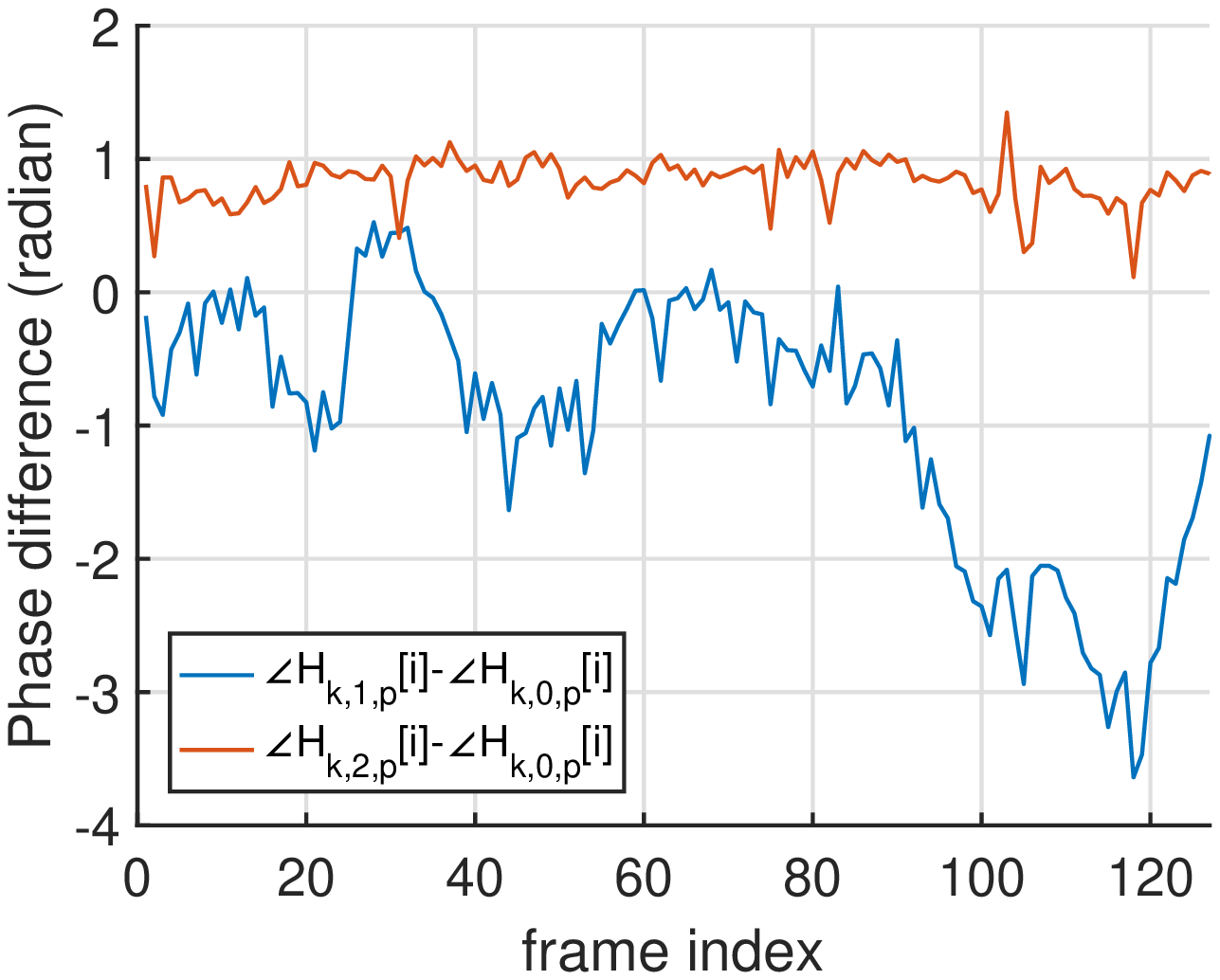}
		\caption{Human movement}
		\label{fig:phase_motion_env}
	\end{subfigure}
	\caption{CSI phase difference between antennas variation over time}
	\label{fig:antenna_phase_test}
	\end{center}
\end{figure}

Multiple antennas at WiFi transceivers are also exploited in this paper to make the phase information of CSI estimate much more useful for presence detection. As WiFi devices use a single oscillator for RF circuitry corresponding to different antennas, the CFO, if present, is common to all inputs at different receive antennas. Similarly, sampling is driven by a single clock, hence STO is also identical for all inputs at different receive antennas. Thus instead of using the raw CSI phase measurement, one can use phase differences between receive antennas to remove phase variation due to CFO and STO~\cite{li2018training-free, qian2018pads, soltanaghaei2017peripheral, Wang2017Fall}. While such processing has no effect on digital communication performance (e.g., it does not correct residual CFO/STO for each receive chain for symbol detection), it cleans up the phase information when phase variation due to human movement is of interest.
An example of phase differences 
is given in Fig.~\ref{fig:antenna_phase_test} with $N_r=3$ where the CSI phase from the first antenna serves as a reference. 
Clearly, phase differences stay relatively stable in a human-free environment whereas  human motion introduces significant fluctuation to the relative phases across three receive antennas.


Figs.~\ref{fig:csi_amplitude_test} and \ref{fig:antenna_phase_test} indicate that both magnitude and phase of estimated CSI contain rich information about human motion. 
While in theory, deep learning trained using labeled data appears to be a straightforward exercise, 
the challenge is that for presence detection, there is no clearly defined human motion that one tries to detect. As such, collecting labeled training data with human presence needs to be carefully addressed along with the design of the learning system for presence detection.

%% file: modules/system_design.tex
\begin{figure*}[!t]
	\centering
	\includegraphics[width=0.80\textwidth, height=0.15\textheight]{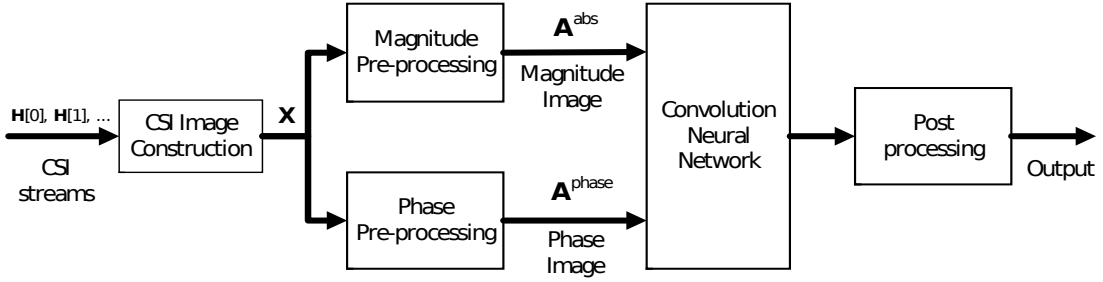}
	\caption{System flowgraph}
	\label{fig:system_structure}
\end{figure*}
\begin{figure*}[!t]
	\centering
	\includegraphics[width=0.75\textwidth, height=0.25\textheight]{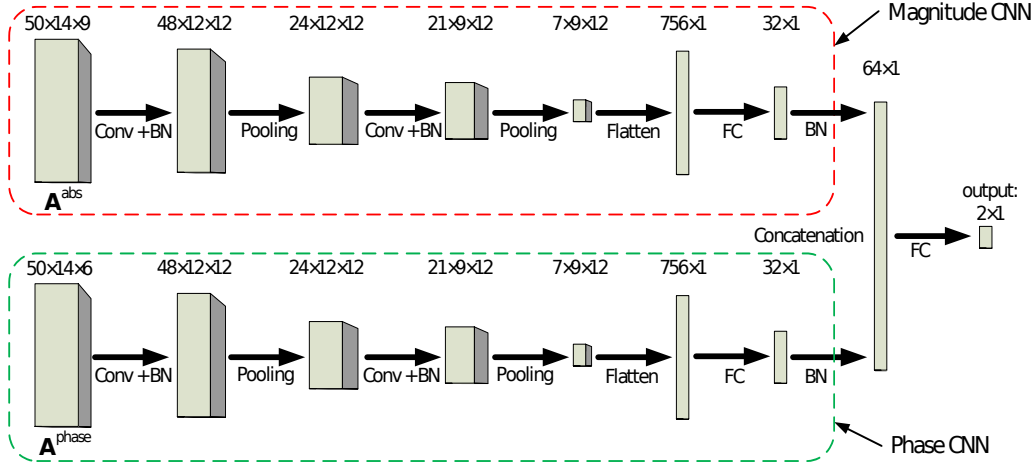}
	\caption{Architecture of the proposed CNN}
	\label{fig:cnn_structure}
\end{figure*}

A high level description of the proposed system is depicted in Fig.~\ref{fig:system_structure}. Consecutive CSIs, extracted using the Atheros-CSI-Tool~\cite{csi-tool}, are first arranged into CSI magnitude and phase images. They are pre-processed separately and fed into the CNN block comprised of two parallel CNNs - one for magnitude and the other for phase - followed by fully connected (FC) layers (c.f. Fig.~\ref{fig:cnn_structure} and Section~\ref{sec:CNN}). The post-processing block accumulates instantaneous detection results provided by CNN and output the final presence detection for a specified time resolution. 

\subsection{Input Pre-processing}
Recall that $\mathbf{H}[i]$ is an $N_{sc}\times N_r \times N_t$ array consisting of MIMO channel matrices across all subcarriers for the $i$-th frame. Instantaneous motion detection is based on  $\mathbf{H}[i]$ collected over $I$ consecutive frames, denoted as $\mathbf{H}[0],\cdots, \mathbf{H}[I-1]$. Here we assume without loss of generality (WLOG) the first CSI array has frame index $0$. For each $\mathbf{H}[i]$, we select $N_{f}$ evenly spaced subcarriers out of $N_{sc}$ subcarriers, resulting in  $\tilde{\mathbf{H}}[i]$ with size $N_f \times N_r \times N_t$. Down selection of subcarriers significantly reduces the data dimension yet does not have any negative effect in sensing performance. This is because the carrier spacing in WiFi signals ($312.5$kHz) is much smaller than the coherent bandwidth in a typical indoor (i.e. low mobility) environment, thus the behaviors of subcarriers that are immediate neighbors closely track each other with or without human motion. The resulting $\tilde{\mathbf{H}}[i]$ are subsequently stacked up along the temporal domain to form a $4$-D array $\mathbf{X}$ of size $I\times N_f \times N_r\times N_t$. The magnitude and phase information are then extracted from $\mathbf{X}$ prior to independent pre-processing. 
 
\subsubsection{CSI magnitude}
We reshape the $4$-D array $|\mathbf{X}|$ into a $3$-D array by combining the last two spatial dimensions, i.e., channel matrix for each subcarrier is flattened into a $1$-D array. The obtained array, denoted by $\mathbf{X}^{\text{abs}}$, is of size ${I}\times N_{f}\times(N_{r}N_{t})$.

Pre-processing $\mathbf{X}^{\text{abs}}$ involves normalization and transformation. 
Normalization is done to remove dependence of the absolute CSI magnitude on various environment parameters that are irrelevant to presence detection. For example, the dynamic range of $\mathbf{X}^{\text{abs}}$ is highly dependent on the distance between the transmitter and receiver and the existence of line of sight transmission. 
While various normalization methods can be used, we find through extensive experiments the following offers the most robust performance: for $i=0,\cdots, I-1$,
 \begin{equation}\label{eq:time_div}
 \tilde{\mathbf{X}}^{\text{abs}}_{i,:,:} = \mathbf{X}^{\text{abs}}_{i,:,:}./\mathbf{X}^{\text{abs}}_{0,:,:},
 \end{equation}
where $./$ denotes element-wise division. Note that $i$ indexes OFDM frame, thus the normalization is done with respect to the first OFDM frame within the $I$ frames contained in $\mathbf{X}^{\text{abs}}$. 

Subsequently, a 2-D DFT is applied to $\tilde{\mathbf{X}}^{\text{abs}}_{:,:,j}$
along the temporal (frame) and frequency (subcarrier) dimensions, resulting in the output array for each transceiver antenna pair:
\[
\tilde{\mathbf{X}}^{\text{abs-fft}}_{:,;,j}=\mathcal{F} \left(\tilde{\mathbf{X}}^{\text{abs}}_{:,:,j}\right).
\]
Here the DFT output is properly shifted so that zero frequency is at the center of the array. The use of 2-D DFT serves two purposes. First, human motion induced temporal variation of CSI is continuous in nature. As such, it results in dispersion in the lower frequency region along the temporal dimension. This is in contrast to hardware impairment and channel estimation error when sudden change of CSI may be observed irrespective of human presence. Therefore, high frequency change can be removed by simple cropping of the DFT output along the temporal dimension around zero frequency:
\begin{equation}\label{eq:shrink_image_magnitude}
	 \tilde{\mathbf{X}}^{\text{abs-fft-crop}}_{i,:, :}=\left|\tilde{\mathbf{X}}^{\text{abs-fft}}_{\frac{I-T}{2}+i,:, :}\right|,
 \end{equation}
 where $i=0,\ldots,T-1$, and $T$ is the cropping window size. Here we assume WLOG that both $I$ and $T$ are even numbers. Cropping also significantly reduces the image size, leading to faster learning and reduced storage requirement. This makes the learning suitable to be implemented on edge devices instead of having to resort to cloud services. Note that further reduction of image size can be achieved by utilizing the conjugate symmetry of the $2$-D FFT due to the fact that input to the FFT is real-valued (i.e., magnitude of CSI arrays). 

\begin{figure}[!t]
	\begin{subfigure}[b]{0.24\textwidth}
		\centering
		\includegraphics[width=1\textwidth]{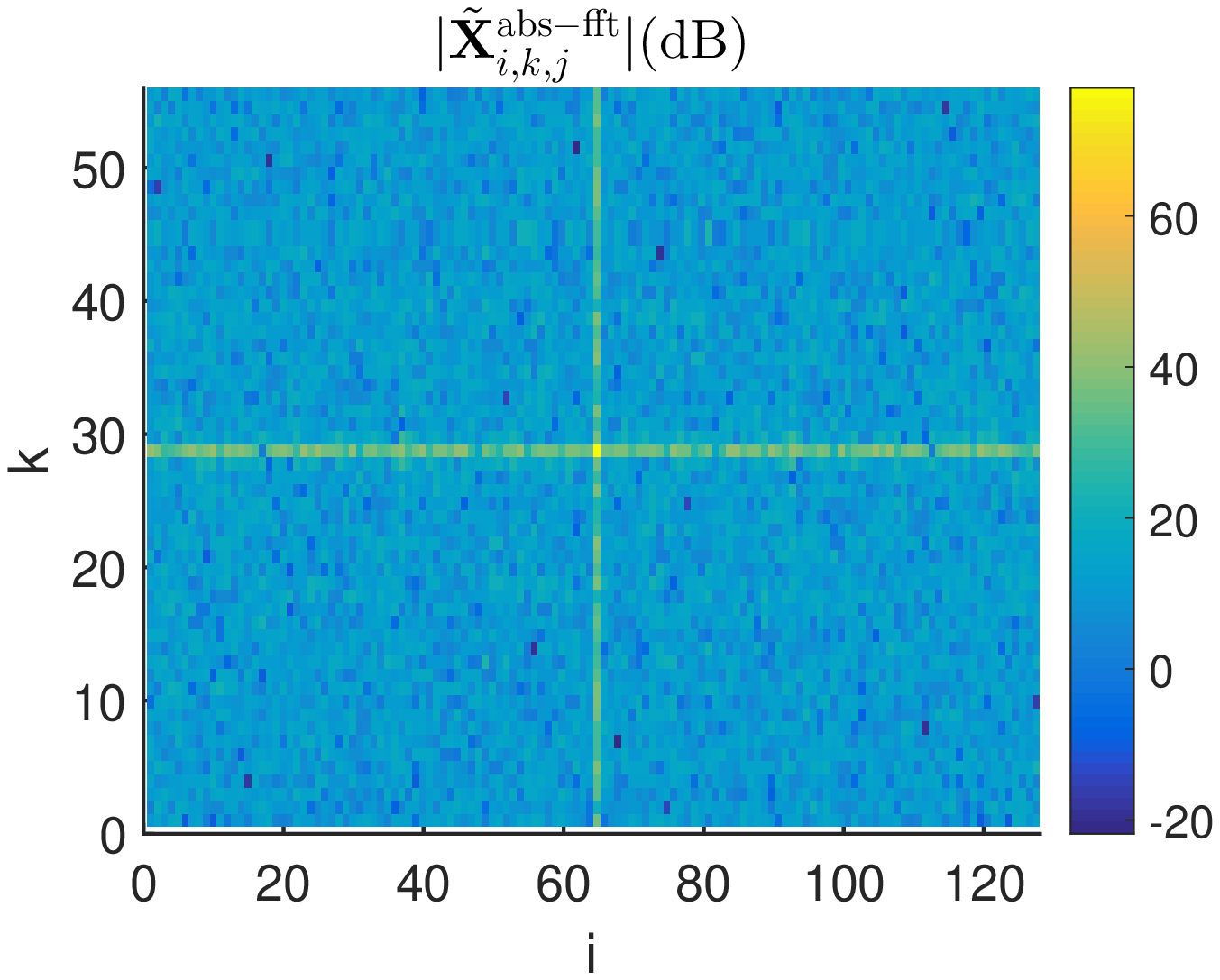}
		\caption{Human-free environment}
		\label{fig:2d_fft_abs_static}
	\end{subfigure}
	\begin{subfigure}[b]{0.24\textwidth}
		\centering
		\includegraphics[width=1\textwidth]{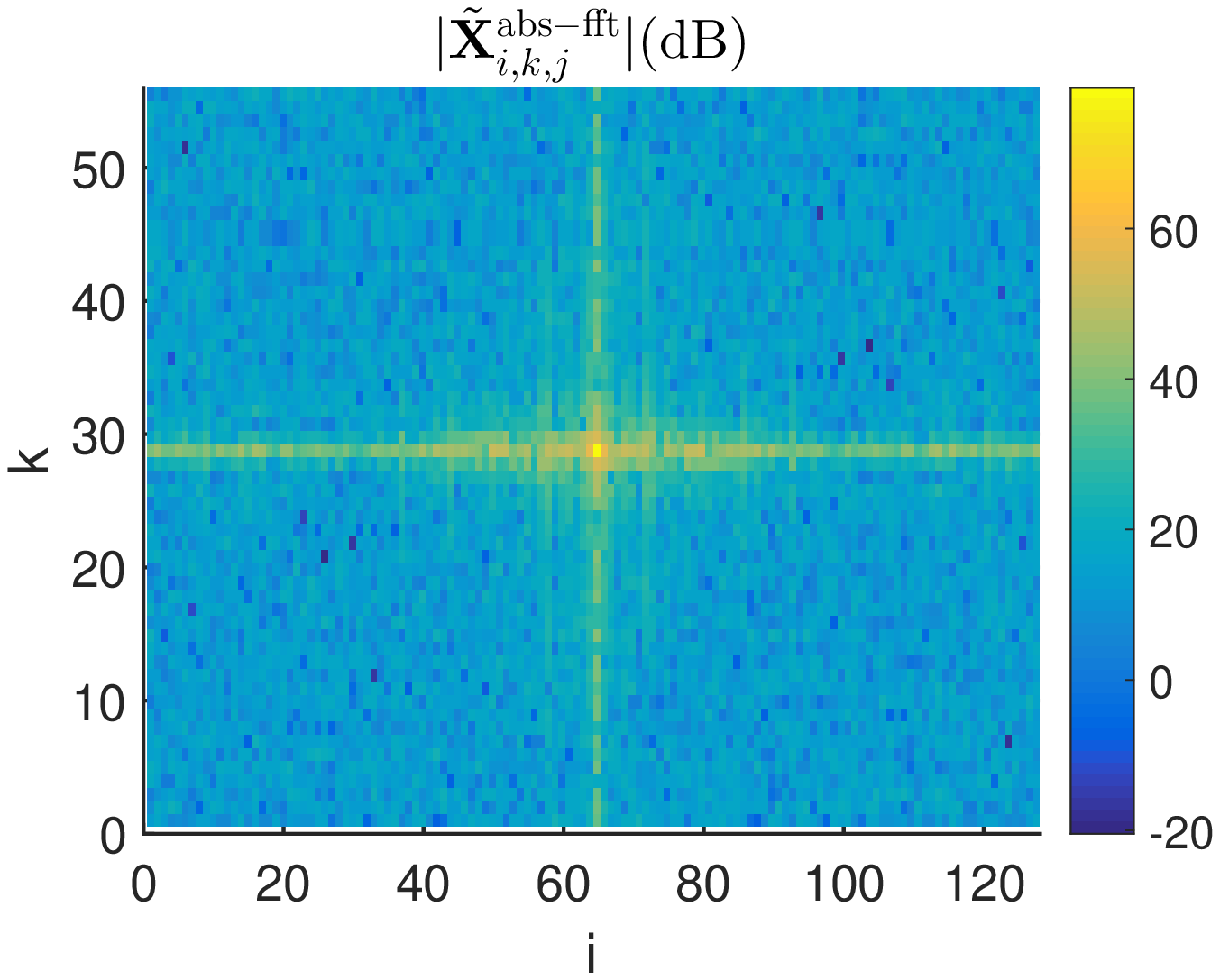}
		\caption{Human movement}
		\label{fig:2d_fft_abs_motion}
	\end{subfigure}
 	\caption{2D DFT of CSI magnitude along frame and subcarrier}
 	\label{fig:2d_fft_abs}
 \end{figure}

Another reason of using 2-D DFT is its ability to localize motion related CSI variation. While temporal variation in $\tilde{\mathbf{X}}^{\text{abs}}_{:,:,j}$ is exhibited for the entire $I$ frames,  2-D DFT concentrates such variation into the low frequency region. This is particularly suitable for CNN given its ability to build discriminating ability on local features. Figs.~\ref{fig:2d_fft_abs_static} and \ref{fig:2d_fft_abs_motion} provide a sample of $\left|\tilde{\mathbf{X}}^{\text{abs-fft}}\right|$ collected in the same room without and with human motions. One can see that, in the temporal dimension, the $2$-D DFT using data collected in an empty room is dominated by the DC component. With human movement, there is clearly dispersion at low frequency region in the temporal (horizontal) dimension.

 \begin{figure}[!t]
	\begin{subfigure}[b]{0.24\textwidth}
		\centering
		\includegraphics[width=1\textwidth]{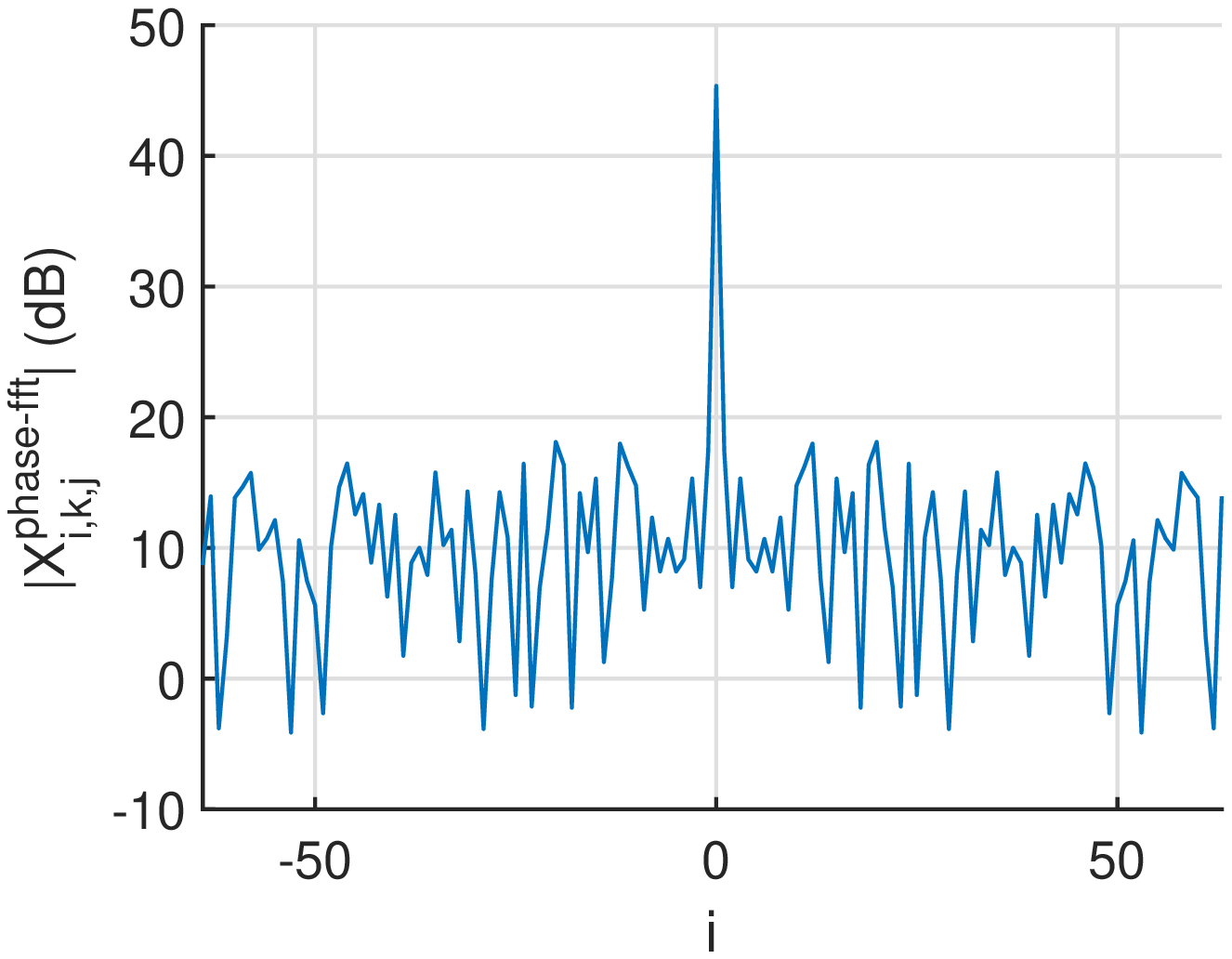}
		\caption{Human-free environment}
		\label{fig:1d_fft_phase_diff_static}
	\end{subfigure}
	\begin{subfigure}[b]{0.24\textwidth}
		\centering
		\includegraphics[width=1\textwidth]{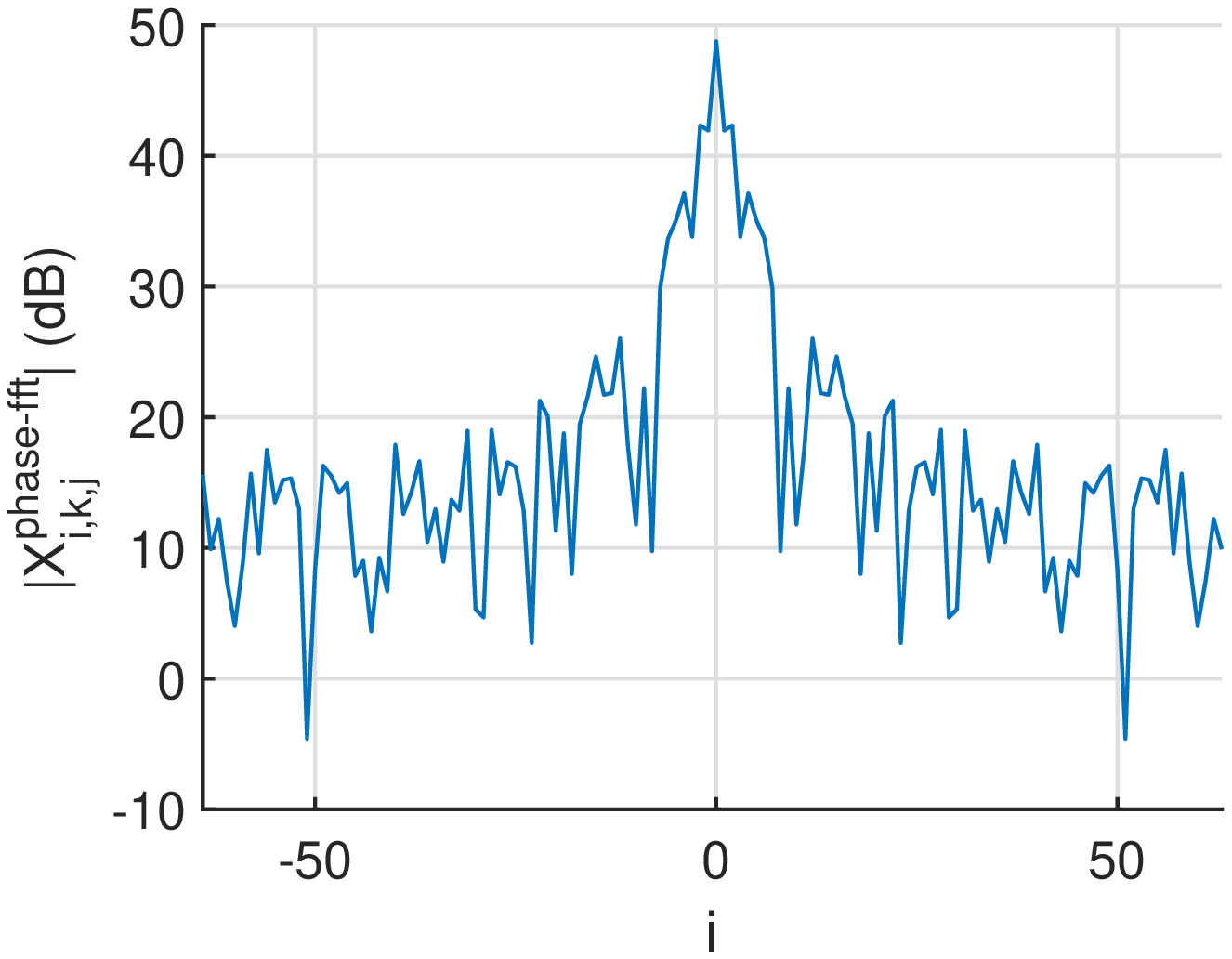}
		\caption{Human movement}
		\label{fig:1d_fft_phase_diff_motion}
	\end{subfigure}
	\caption{DFT of CSI phase difference at a fixed subcarrier}
	\label{fig:1d_fft_phase_diff}
\end{figure}

\begin{figure}[!t]
	\begin{subfigure}[b]{0.24\textwidth}
		\centering
		\includegraphics[width=1\textwidth]{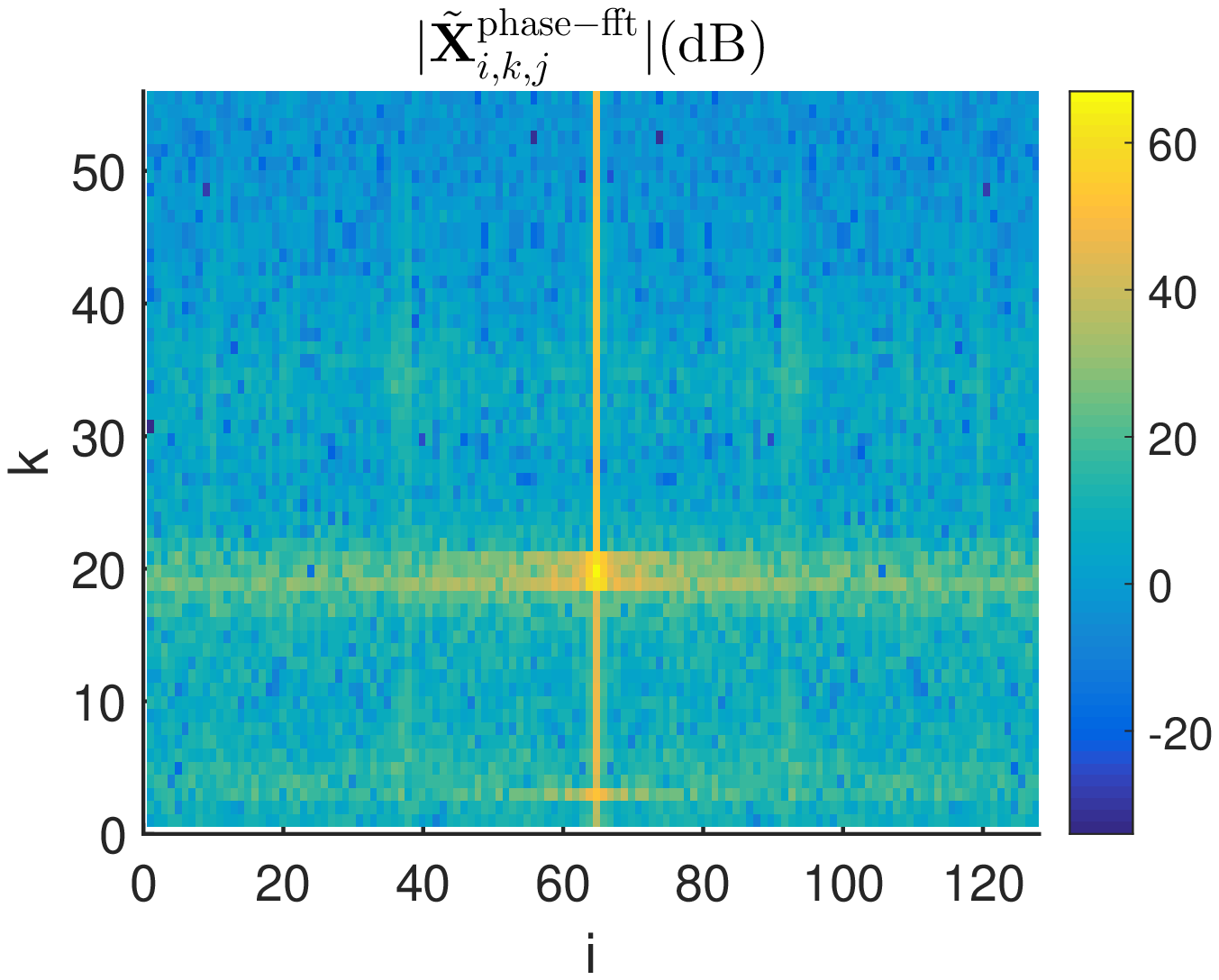}
		\caption{Human-free environment}
		\label{fig:1d_fft_phase_diff_static_image}
	\end{subfigure}
	\begin{subfigure}[b]{0.24\textwidth}
		\centering
		\includegraphics[width=1\textwidth]{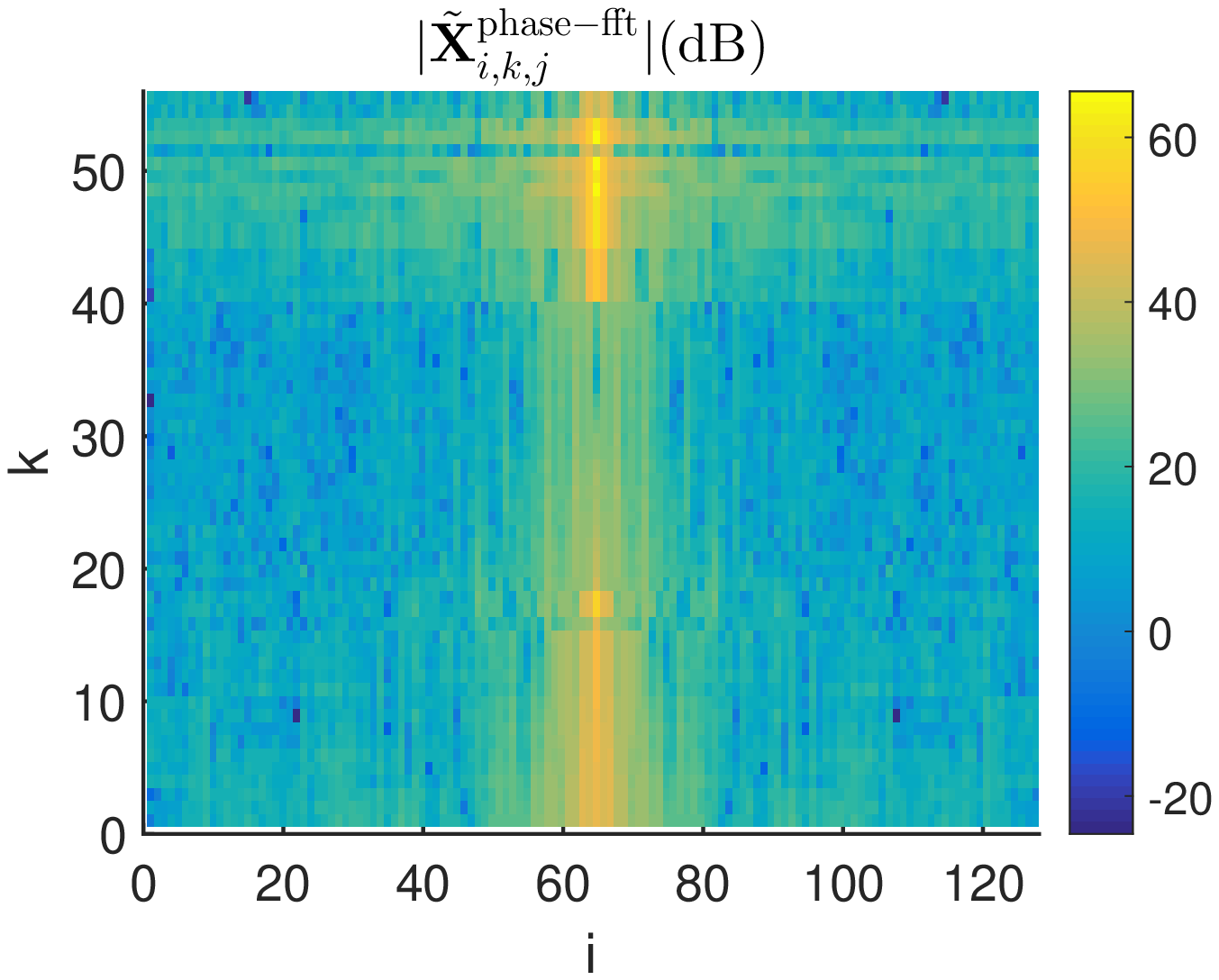}
		\caption{Human movement}
		\label{fig:1d_fft_phase_diff_motion_image}
	\end{subfigure}
	\caption{DFT of CSI phase difference along frame at all subcarriers}
	\label{fig:1d_fft_phase_diff_image}
\end{figure}

\subsubsection{CSI phase}
Even with a completely static environment, the estimated CSI phase will undergo variation (e.g., from residual CFO and STO) which may lead to abrupt changes within $(-\pi, \pi]$. 
This can be partially resolved by phase unwrapping which removes such abrupt changes. However, phase unwrapping does not remove phase variation introduced by any residual CFO and STO offset, but merely correct discontinuous phase jumps. Thus CSI phases are often discarded for WiFi sensing \cite{palipana2016nonlinear, gong2016adaptive, zhu2017r} because of this ``noisy" nature. 

A simple pre-processing that computes phase difference with respect to a reference receive antenna can largely mitigate this problem due to the fact that CFO and STO are common to all receive antennas (see Section~\ref{sec:motion}). 
Denote by $\mathbf{X^{\text{phase}}}$ the phase difference between $\tilde{\mathbf{H}}_{k,q,p}[i]$ for different $q$
\begin{equation} \label{eq:phase_offset_calc}
\mathbf{X}^{\text{phase}}_{i,:,q-1,:} = \angle(\tilde{\mathbf{H}}_{:,q,:}[i]./\tilde{\mathbf{H}}_{:,0,:}[i]),
\end{equation}
where $q=1,\ldots,N_{r}-1$. 
The last two spatial dimensions of $\mathbf{X}^{\text{phase}}$ are then flattened into one dimension and phase is unwrapped along the time axis to remove discontinuity at boundary points $-\pi$ and $\pi$. The obtained result is denoted by $\tilde{\mathbf{X}}^{\text{phase}}\in\mathbb{R}^{I\times{N_{f}}\times{(N_{r}-1){N_{t}}}}$.
Different from CSI magnitude, only $1$-D DFT along the temporal dimension is performed on $\tilde{\mathbf{X}}^{\text{phase}}$ to get $\tilde{\mathbf{X}}^{\text{phase-fft}}_{:,k,j}=\mathcal{F}\left[\tilde{\mathbf{X}}^{\text{phase}}_{:,k,j}\right]$ since phase unwrapping weakens relation of CSI phase across different subcarriers.
An example of $\left|\tilde{\mathbf{X}}^{\text{phase-fft}}\right|$ is given in Figs.~\ref{fig:1d_fft_phase_diff} and \ref{fig:1d_fft_phase_diff_image} where significantly increased dispersion of the DFT output along the temporal dimension can be observed with human movement.

The following steps are similar to how we obtain $\tilde{\mathbf{X}}^{\text{abs-fft-crop}}$, where we shift the zero frequency component to the center and crop out the high frequency components in the temporal domain, leading to the following CSI phase information
\begin{equation}\label{eq:shrink_image_phase}
\tilde{\mathbf{X}}^{\text{phase-fft-crop}}_{i,:,:}=\left|\tilde{\mathbf{X}}^{\text{phase-fft}}_{\frac{I-T}{2}+i,:,:}\right|,
\end{equation}
where $T$ is chosen to be the same as that in \eqref{eq:shrink_image_magnitude}.

\subsubsection{Image Normalization}
DFT typically results in increased dynamic range of $\tilde{\mathbf{X}}^{\text{abs-fft-crop}}$ and $\tilde{\mathbf{X}}^{\text{phase-fft-crop}}$. Elements with low intensity are easily overwhelmed by those with large values. The logarithmic operator $y=\log_{10}(x+1)$ is applied to each element in both images~\cite{image_process_book} to reduce such disparity.  
The final input to the two parallel CNNs are
\begin{equation}
\begin{aligned}
\mathbf{A}^{\text{abs}}&=\log_{10}(\tilde{\mathbf{X}}^{\text{abs-fft-crop}}+1), \\
\mathbf{A}^{\text{phase}}&=\log_{10}(\tilde{\mathbf{X}}^{\text{phase-fft-crop}}+1). \\
\end{aligned}\label{eq:input}
\end{equation}

\subsection{Architecture of CNN \label{sec:CNN}}

The architecture of the proposed CNN is shown in Fig.~\ref{fig:cnn_structure}. Magnitude and phase images in (\ref{eq:input}) are fed into two parallel CNNs which share the same structure. The output of the two CNNs are then concatenated and fed to FC layers. 

The building blocks of the proposed CNN are similar to  AlexNet~\cite{krizhevsky2012imagenet}. Each of the two parallel CNNs consists of two convolution (Conv) layers without padding. 
Each Conv layer is followed by an average pooling layer~\cite{lecun1998gradient} to reduce the output dimension.  The multi-dimensional output of the last Conv layer is flattened into vectors and subsequently fed into an FC layer. Batch Normalization (BN)~\cite{ioffe2015batch} is added after each layer that has trainable parameters to speed up training and make the model more robust against variations in outputs from previous layers. Two activation functions are used - rectified linear unit (ReLU) for the hidden layers and softmax for the output layer. We note that with presence detection, the number of classes is $2$, i.e., it is a binary classification problem. Therefore, a sigmoid function can be used instead of softmax for the output layer. However, our experiment indicates slightly more robust classification performance using softmax - this can perhaps be attributed to the difference in weight and bias terms between the two: softmax employs two independent sets of weight vectors and biases for the two neurons whereas the sigmoid function has a single input to the neuron at the output layer. While mathematically one can show equivalence between the two for binary classification by finding the corresponding parameters, learning such parameters through training may yield some performance difference.

In the training phase, cross-entropy is chosen to be the loss function and Adam optimizer~\cite{kingma2014adam} is used to update weights during backpropagation. To prevent overfitting, both $l_2$ regularization and dropout layers with dropout probability $0.5$ are added for each fully-connected hidden layer.

\subsection{Post processing}

The design of the post processing block is closely tied with how data collection is conducted. As alluded in the introduction, presence detection differs with detection of certain activities in that one is not looking for a certain activity pattern but rather, whether a room is being occupied or not, assuming that occupants are not completely still for extended periods of time. As such, there are two different ways of collecting training data for the occupant state: one is to collect CSI for the entire duration when occupants are present; an alternative way is to collect CSI only when occupants are moving. While in theory the former seems to be a natural choice - what we try to detect is the presence or absence of humans in a room - doing so leads to significantly high false alarm rate regardless of how many training data are collected. The reason is  quite simple: collecting training CSI data when humans are present will include many instances when humans are completely still. Such CSI samples, {\em albeit} scattered throughout the measurement data (i.e., not for extended period of time), are indistinguishable with that of an empty room. In essence, the training data corresponding to human present are polluted with a large number of data samples that are similar to that training data without human presence. 

We elect to use training data corresponding to the CSI instances when there are detectable human movements in the room. While this leads to `missed detection' corresponding to instances when the occupants are still, simple post processing can be done after the CNN block with tunable parameters such as the resolution with which the presence detection is desired. In short, the training is done so that the CNN attempts to reliably detect human motion of any kinds; complete still human presence thus is likely to be classified as the negative state. Post-processing then applies some averaging operation within a time window, whose duration corresponds to some desired time resolution, for presence detection. This is sufficient in practice since with a truly empty room, the CNN output should contain negative outputs whereas with human present, the output should have significant portion of positive outputs.



%% file: modules/experiment.tex
 This section describes the experiment setup where COTS WiFi cards are used to collect WiFi CSI in an indoor environment. Data collection is explained in detail and the presence detection result is compared to that using PIR sensors.
 
 \subsection{Experiment setup}
 Our WiFi system consists of a laptop (Thinkpad T$410$) as WiFi access point (AP) and a desktop (Dell OptiPlex $7010$) as WiFi client. Atheros 802.11n WiFi chipset, AR9580, and Ubuntu 14.04 LTS with built-in Atheros-CSI-Tool~\cite{csi-tool} are installed on both computers. The AP sends packets at the rate of $100$ pkts/s. The client records CSIs using Atheros-CSI-Tool,  and the CSI sampling interval is roughly $10$ms. 
 With $N_r=3$ receive antennas, $N_t=3$ transmit antennas, and $N_{sc}=56$ subcarriers in a $20$MHz channel operating at channel $6$ in the $2.45$GHz band \cite{ieee80211n},  each CSI instance $\mathbf{H}[i]$ is a $56\times 3\times 3$ complex valued array. Down-selecting to $N_f=14$ evenly spaced subcarriers, the resulting $\tilde{\mathbf{H}}[i]$ is of dimension $14\times 3\times 3$.
 
\begin{figure}[!t]
    \centering
    \includegraphics[scale=0.72]{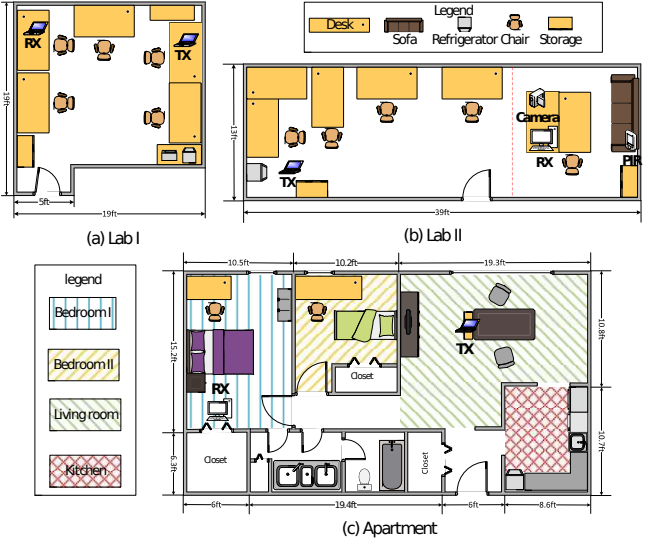} 
    \caption{Indoor space layout}
    \label{fig:floor_plan}
\end{figure}

\begin{figure}[!t]
\begin{tabular}{ccc}
 		\includegraphics[width=0.14\textwidth, height=0.1\textwidth]{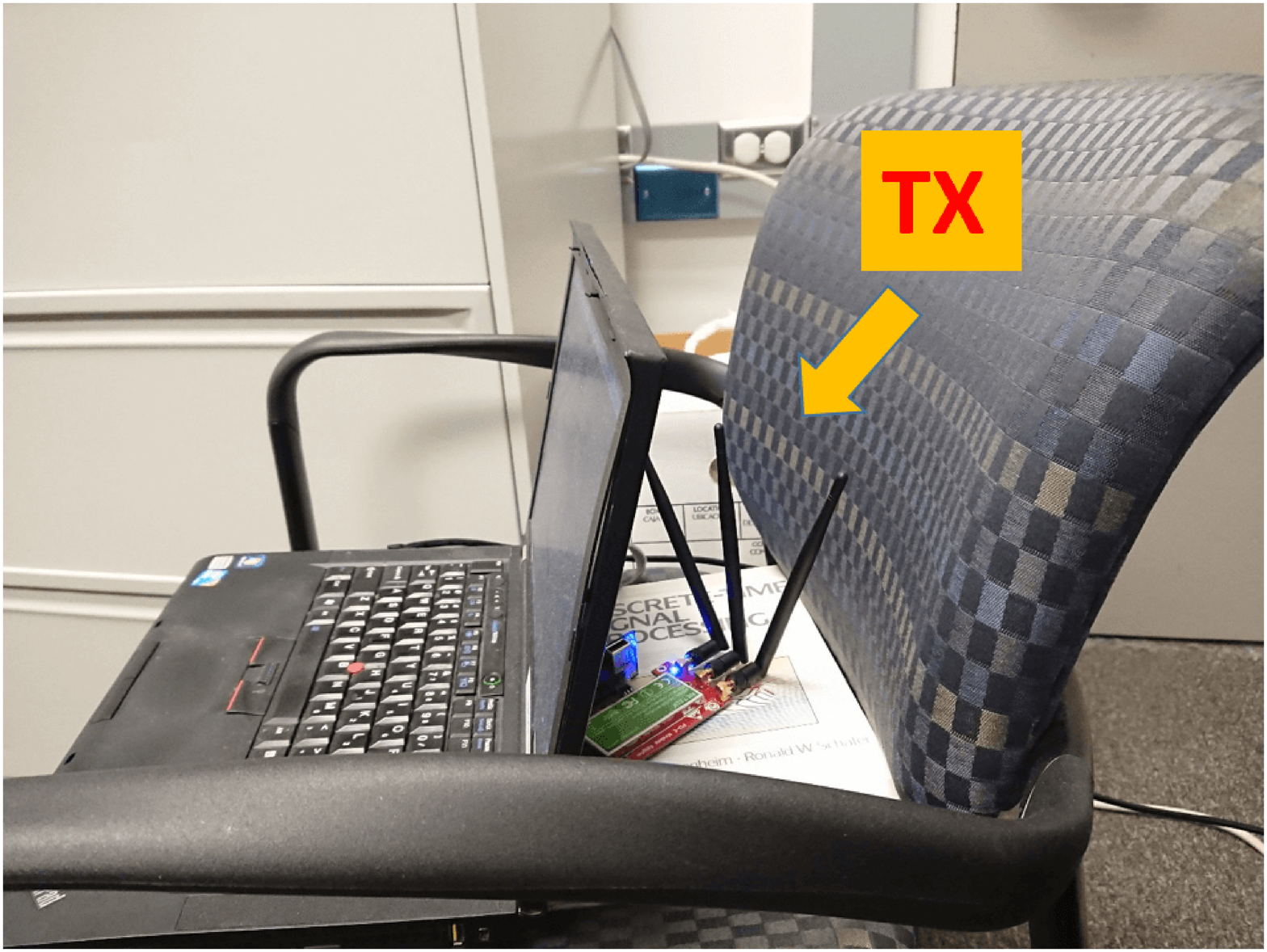}
&		\includegraphics[width=0.14\textwidth, height=0.1\textwidth]{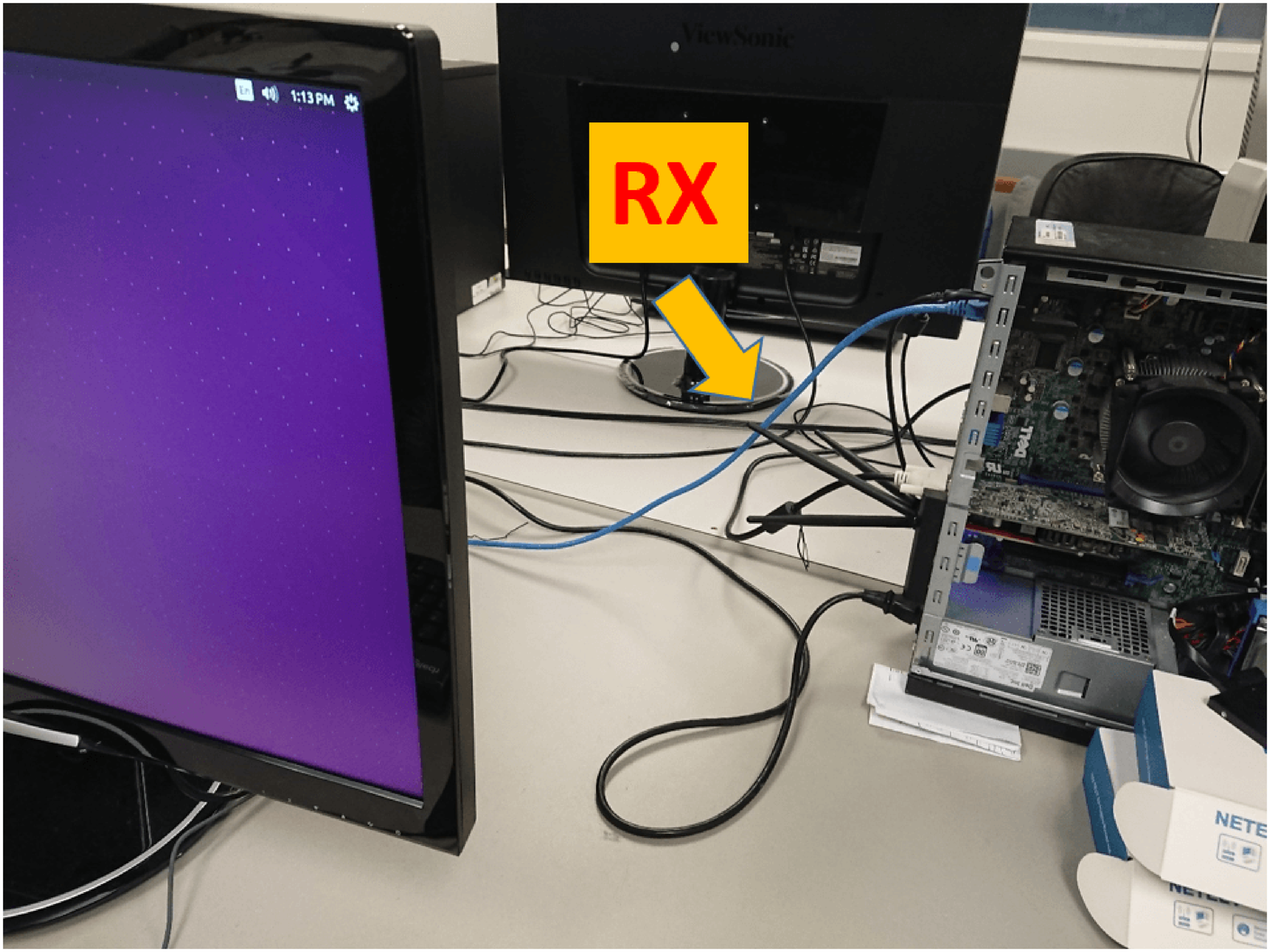}
&	\includegraphics[width=0.14\textwidth, height=0.1\textwidth]{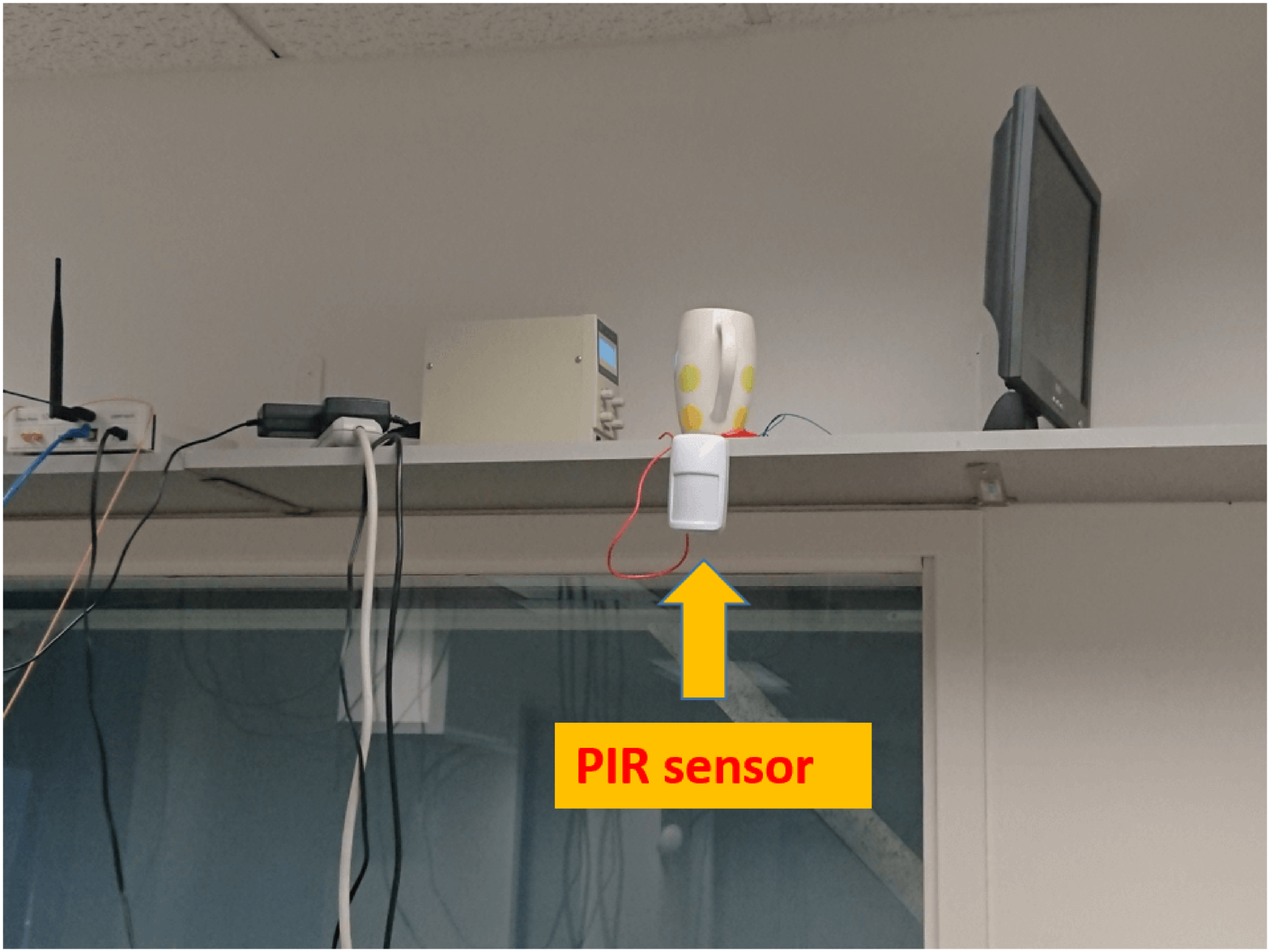}\\
(a) Transmitter & (b) Receiver & (c) PIR sensor
\end{tabular}
	\caption{Device setup}
	\label{fig:tx_rx_setup}
\end{figure}

 The indoor environment in which both training data collection and testing are done is sketched in Fig.~\ref{fig:floor_plan}. 
 Three different environments are used: two labs of different size and layout and a typical two-bedroom apartment in a four-story apartment building. 
 In the lab environment, there are multiple monitors/laptops on desks and more chairs on the floor; these are not drawn in the figure and their positions may change in different days. The transmit antennas are placed behind a laptop and the receive antenna array is surrounded by a lot of other computers as shown in Fig.~\ref{fig:tx_rx_setup}. Therefore, there is no strong line of sight component between the transmitter and the receiver. For the two-bedroom apartment, only large furniture such as beds and tables are sketched in the plot.
 
\subsection{Data collection}
For the image input to the CNN, we choose $I=128$ consecutive CSI instances, which lasts for around $1.27$s. This is chosen since one second is sufficiently long for any detectable human motion to induce temporal CSI variation. 
A CSI image is only used (i.e., considered a valid sample) if it satisfies the following two conditions: 1) Every entry of $\left|\mathbf{X}\right|$ is non-zero. This is imposed to remove erroneous CSI estimate - occasionally zero entries will show up in recorded CSI series, potentially due to hardware/firmware problems.
2) The time difference between the last and the first frame lies within $1.27\pm0.064$s. WiFi scheduling may lead to different frame lengths hence excessively long interval between two CSI estimates.
In the experiment, the cropping window size is chosen to be $T=50$, hence $\mathbf{A}^{\text{abs}}$ and $\mathbf{A}^{\text{phase}}$ in \eqref{eq:input} are of size ${50\times14\times9}$ and ${50\times14\times6}$ respectively. 

Data collected in the human-free state is labeled $0$. 
The training data with label $1$ are collected when at least one person is walking randomly in the room. This way, training samples collected when occupants are completely still will not be used. 
Both human-free and motion data are collected on multiple days since the wireless channels are inherently nonstationary. 
This prevents CNN from being tuned to features that are irrelevant to presence detection, e.g., different CFO and STO on different days due to frequency/time drift. Data collection on any given day is also divided into disjoint runs which alternate between human-free and human motion. Finally, training and test data come from completely disjoint days. 


The CSIs were collected during $24$ days at three different locations over a period of $8$ months from 09/14/2019 to 05/17/2020. Data collected during the first three days were from Lab I, days 4-19 were from Lab II, and the last five days were from the apartment, respectively. Spreading out the measurement data over an extended period of time helps us to understand the sensitivity of the learned system to environmental change. There is no deliberate effort to maintain the same furniture arrangement during the measurement duration.

The proposed CNN is built under Keras with Tensorflow as backend~\cite{chollet2015keras}. Training and off-line testing described in Section~\ref{exp::part1} and Section~\ref{exp::perf_cmp} are performed on a Linux server (Dell PowerEdge R$730$) with one E$5$-$2650$ v$4$ CPU and $128$GB of RAM. On-line detection described in Section~\ref{exp::part2} is run on the WiFi receiver (Dell desktop) with one i$7$-3770 CPU and $8$GB of RAM.


\subsection{Motion Detection}\label{exp::part1}
We first evaluate motion detection using the proposed CNN without post-processing, i.e., CNN is trained to classify input CSI images according their labels: $0$ for human free data whereas $1$ for data with someone randomly walking around. 

The CNN with $55078$ parameters is trained using data from days $9-14$ in Lab II which span two weeks. The number of training data in each class is summarized in Table~\ref{table:training_data_composition}. Both labels have  roughly $40000$ images which correspond to $30-$min data collected each day. The learned model, denoted by model I, is trained for 10 epochs, lasting for a total of $156.30$ seconds. Test data are from days $1-5, 15-16$ and $20-24$; motion data collected on those days have  similar motion types as the training data, i.e., they were collected when human introduced large motions such as walking, sitting down, and standing up. Test data were collected at both Lab II (i.e., same as training) and at Lab I and apartment (i.e., different from training). 

\subsubsection{Testing in the same environment} \label{exp::test_same_environment}

\begin{table}[!t]
	\renewcommand{\arraystretch}{1.5}
	\caption{Training set composition}
	\centering
	\begin{tabular}{ | c| c| c |c|c|}
		\hline
		\multirow{2}{*}{Model name} & \multicolumn{2}{c|}{human free} & \multicolumn{2}{c|}{human motion} \\ \cline{2-5}
		& days & size & days & size \\ \hline
		model I & $9-14$ & $39866$ & $9-14$ & $41276$ \\ \hline
		model II & $3$, $9-14$ & $24642$& $9-14$ & $20383$ \\ \hline
		model III & $9-14$ & $19617$ & $20, 22$, $9-14$ & $23831$ \\ \hline 
		model IV & $8$, $9-14$ & 50753 & $9-14$ & $41276$ \\ \hline
	\end{tabular}	
	\label{table:training_data_composition}
\end{table}

\begin{figure}[!t]
	\centering
	\includegraphics[scale=0.4]{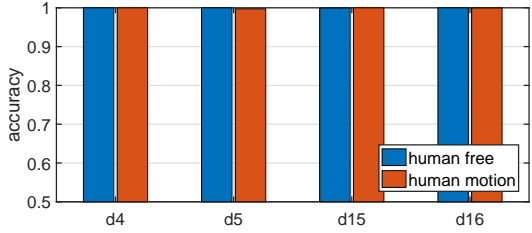}
\caption{Performance in Lab II on different test days}
\label{fig:performance_lab2}
\end{figure}

Model I is first tested using Lab II data collected on days $4,5,15,16$.
There are roughly $5000$ CSI images for each label per day corresponding to about half an hour measurements. 
The results are shown in Fig.~\ref{fig:performance_lab2}.
The performance for both labels is quite consistent (all close to $100\%$) - we note that days $4$ and $5$ were collected one month earlier than the training data, thus lab settings, e.g., the transceiver placement and the location/number of surrounding objects were quite different. The proposed CNN is therefore quite robust to the environment changes over time. 

\subsubsection{Testing in a different environment}\label{exp::test_different_environment}
\begin{figure}[!t]
	\centering
	\includegraphics[scale=0.4]{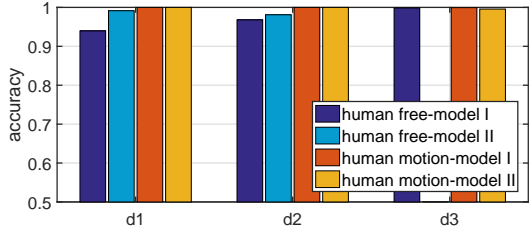}
    \caption{Performance in Lab I on different test days}
\label{fig:performance_lab1}
\end{figure}

The performance of model I in Lab I is given in Fig.~\ref{fig:performance_lab1}. Even though sensitivity to human motions remains high, there is noticeable increase in the false alarm rate. For example, the false alarm rates in day $1$ and day $2$ are $6.05\%$ and $3.21\%$ respectively. A simple remedy is to include human free data collected from Lab I in the training set. This leads to Model II in Table~\ref{table:training_data_composition} where human free data on day $3$ which comes from 30-min runs are combined with $40000$ randomly chosen samples from the previous training set. From Fig.~\ref{fig:performance_lab1}, the new model exhibits noticeable improvement in the first two days' false alarm rate without introducing degradation in the human detection rate. 

\begin{figure}[!t]
	\centering
	\includegraphics[scale=0.35]{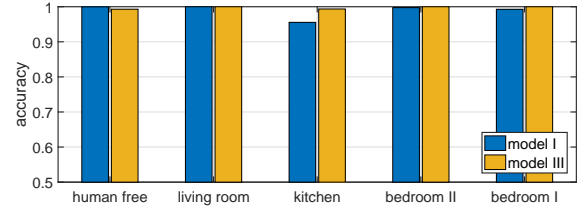}
    \caption{Performance in the apartment at different locations}
\label{fig:performance_apt}
\end{figure}

Model I is further evaluated using data from days $20-24$ in a completely different and more complex apartment environment. Different from the above single-room tests, results are now categorized according to where the motions took place. Specifically, human motions can happen  in four different rooms: living room, kitchen, bedroom I and bedroom II. The accuracy given in Fig.~\ref{fig:performance_apt} is averaged over all the test days. Without any data from the unseen environment, the model still has high detection rates (close to $100\%$) in three rooms except the kitchen ($95.55\%$) while at the same time has a low false alarm rate ($0.04\%$). Notice that bedroom II has two walls between the transmitter and receiver, indicating that the model has good through the wall detection capability. To further improve the sensitivity in the kitchen, only data for motions in the kitchen from  days $20$ and $22$ are added to the training (day $21$ does not have motion data collected). 
This portion of data comes from runs lasting for $10$ minunites each day. The Model III given in Table~\ref{table:training_data_composition} is trained using the new data together with $40000$ randomly chosen samples from the previous training set collected in the lab environment. The average accuracy for motions in the kitchen excluding days $20-22$  improved from $95.55\%$ to $99.36\%$ with no noticeable performance changes for other rooms in the apartment. 

\subsection{Performance Comparison}\label{exp::perf_cmp}
In this section, we compare the performance of the proposed CNN (Model I) with two recently proposed WiFi based motion detection systems with similar experimental setup: the PADS~\cite{qian2018pads} and R-TTWD~\cite{zhu2017r} systems. PADS leverages the temporal correlation matrix of both CSI amplitudes and phase differences across antenna elements. The first few largest eigenvalues of correlation matrices are adopted as features. R-TTWD exploits CSI amplitude correlation in the frequency domain (i.e.,  across different subcarriers). 
To increase detection of moving humans at different side of a wall, R-TTWD proposes to evaluate means of first-order difference of the first few eigenvectors. Both PADS and R-TTWD use SVM~\cite{libsvm} after feature extraction to find the decision boundary in the feature space for presence detection. 

For a fair comparison,  the same input CSIs are given to three systems with the only difference in the number of subcarriers $N_{f}$. PADS and R-TTWD use $28$ evenly spaced subcarriers instead of $14$ in the proposed CNN to maintain the desired frequency resolution (both systems use all $30$ subcarriers extracted from off-the-shelf NIC, Intel 5300, in the original papers). Besides, all the pre-processing steps are conducted as described in \cite{qian2018pads,zhu2017r}. 

\begin{figure}[!t]
	\begin{subfigure}[b]{0.5\textwidth}
	\centering
	\includegraphics[scale=0.4]{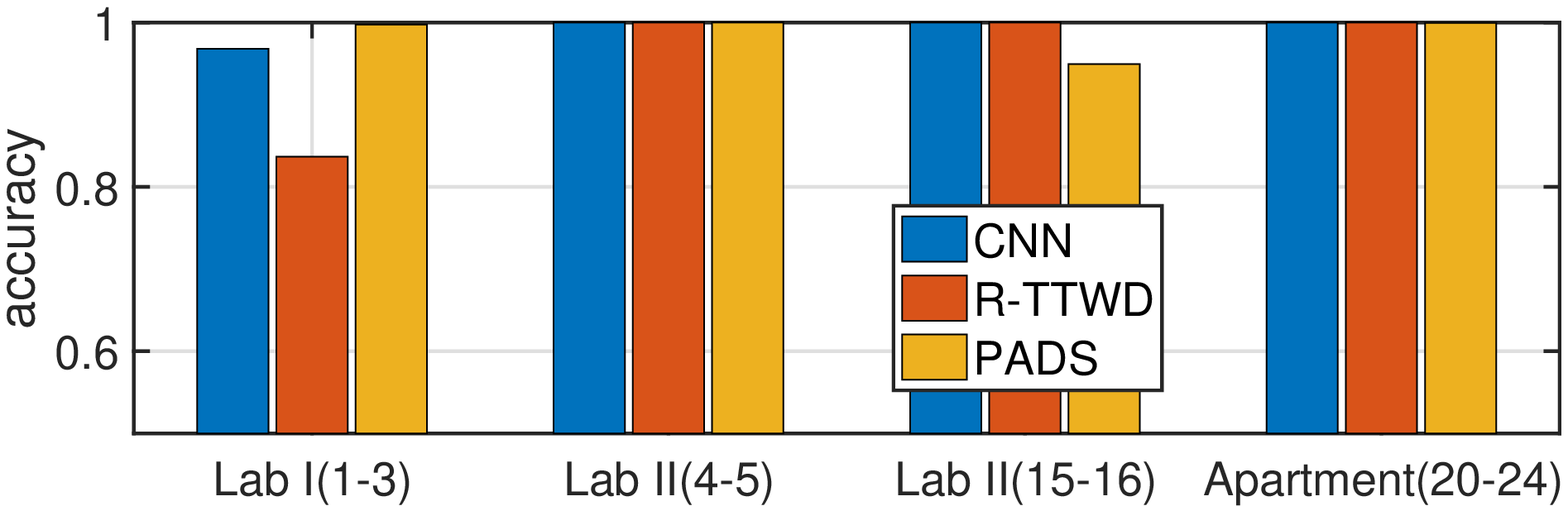}
	\caption{Human free}
	\label{fig:performance_empty_cmp}
	\end{subfigure}
	\begin{subfigure}[b]{0.5\textwidth}
	\centering
	\includegraphics[scale=0.4]{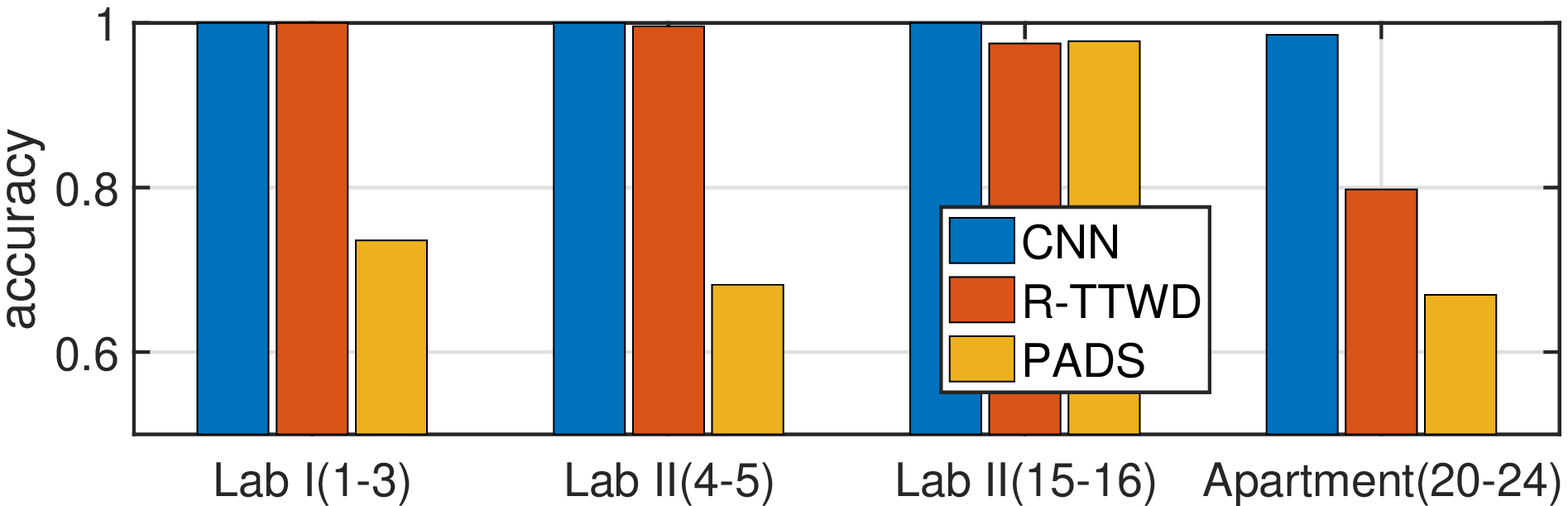}
	\caption{Human motion}
	\label{fig:performance_motion_cmp}
    \end{subfigure}
\caption{Performance comparison}
\label{fig:performance_cmp}
\end{figure}

The average accuracy of the three systems under four different test scenarios are shown in Fig.~\ref{fig:performance_cmp} with test data from days listed in the parentheses of the horizontal axis. In Lab II, i.e., the same environment as the training data, our model and R-TTWD give comparable result. However, when tested in Lab I, R-TTWD has considerably higher false alarm rate compared with  CNN. Performance deteriorates even further in the apartment for R-TTWD as its miss detection rate increases dramatically. 
For the PADS system, even though its false alarm rate remains consistently low (Fig.~\ref{fig:performance_empty_cmp}), its motion detection suffers significantly with environment change (Fig.~\ref{fig:performance_motion_cmp}).

Clearly, compared to both PADS and R-TTWD systems, the proposed learning system performs consistently better and exhibits much more robust performance when tested in a completely new environment.

\subsection{Presence Detection}\label{exp::part2}
Before presenting our presence detection results conducted in Lab II, let us first examine how the CNN trained with walking data performs when more subtle human motion is used for testing. These data are collected on days $6-8$ in Lab II with various small scale motion (turning in chairs, arm waving, etc.). The results are summarized in Table~\ref{table:small_scale_test_result}. Clearly, with training data coming from exclusively random walking for label $1$, the CNN can still reliably detect other motion types.

\begin{table}[!t]
	\renewcommand{\arraystretch}{1.5}
	\caption{Test accuracy for small scale motion}
	\centering
	\begin{tabular}{ | c| c| c | c | c|}
		\hline
		\multirow{2}{*}{Days} & 
		\multicolumn{2}{c|}{human free} &  \multicolumn{2}{c|}{human motion}  \\  \cline{2-5}
		& size & model I & size & model I  \\ \hline
		$6$ & $5396$& $100\%$ &$4992$ & $99.94\%$ \\ \hline
		$7$ & $5447$ & $100\%$ & $5044$ & $98.47\%$ \\ \hline
		$8$ & $10887$ & $99.98\%$ & $4837$ & $96.90\%$ \\ \hline
	\end{tabular}	
	\label{table:small_scale_test_result}
\end{table}

The actual model used for presence detection (model IV in Table~\ref{table:training_data_composition}) is obtained by further augmenting training data with label $0$ data (i.e., human free) collected on day 8. This is done since the output motion probabilities of human-free data collected on day $8$ are closer to $0.5$ than other days. Thus adding these data for CNN training provides more sample diversity. 
Model IV is then deployed at the WiFi receiver, along with post-processing, for real-time presence detection.


As a comparison study, presence detection is conducted concurrently using a PIR sensor. We chose Honeywell DT$8035$~\cite{pir}, a leading edge PIR sensor with a coverage range of $40\text{ft}\times56\text{ft}$ (our lab dimension is $13\text{ft}\times39\text{ft}$). 
A camera is used in the lab to provide ground truth. 
The PIR sensor is mounted on the shelf at one side of the room at a height of $6.8\text{ft}$ (see Fig.~\ref{fig:tx_rx_setup}c). 
Note that DT$8035$ also has a microwave sensor which was disabled for this experiment. 
Throughout the experiment, human activities are restricted to the left side of the room  (left of the red dash line in Fig.~\ref{fig:floor_plan}b) to avoid blind spot of the PIR sensor as its coverage is in a conical shape.

The post-processing is an averaging process on the motion detection outputs of the CNN. Each new CSI instance is used to construct CSI images with the previous $127$ CSI estimates, i.e., a sliding window with step size one is applied to the CSI series. This results in a CNN output at a rate of about one per $10$ms. 
Presence detection output occurs every second. Each second is divided into five subintervals, each of duration $200$ms. For each subinterval, a positive motion detection occurs when at least $10$ CSI images have output label $1$. A positive detection is declared for the one second period if at least three out of the five subintervals have positive motion detection.
Finally, given that the PIR sensor outputs its detection result between $2$ to $5$ times each second, we choose the detection resolution to be $1$ second for both WiFi and PIR: a presence is detected for each second if there is at least one positive detection within the one second period for PIR. 


\subsubsection{False positive test}
The test is done over a $3$ day period (days $17-19$ from 12/30/2019 to 01/01/2020), when Lab II is empty. Results shown in Table~\ref{table:false_alarm_result} are the numbers of one-second intervals in which presence is detected by CNN and PIR sensor. To avoid interruption to normal lab activities, a single test run on certain days can not last for very long. For example, on day $17$, the entire test is broken into three periods, with the shortest one lasting for $20$ minute during lunch break. The entire test lasts for about $46.5$ hours, the proposed system only report false positive four times, yielding a false alarm rate $2.4\times 10^{-5}$. The PIR sensor has zero false alarm rate and the results are comparable given that isolated one second positive can be easily ruled out for occupancy detection.

\begin{table}[!t]
	\renewcommand{\arraystretch}{1.5}
	\caption{False alarm counts (in seconds) in a human-free room}
	\centering
	\begin{tabular}{ | c|c|c| c| c |}
		\hline
		day &  index & duration & CNN & PIR \\ \hline	
		\multirow{3}{*}{17} & 1 & 8hrs & 3s & 0s \\ \cline{2-5}
		& 2 & 20mins & 0s & 0s \\  \cline{2-5}
		& 3 & 9hrs & 1s & 0s \\ \hline
		
		\multirow{2}{*}{18} & 1 & 8hrs & 0s & 0s \\ \cline{2-5}
		& 2 & 9hrs & 0s & 0s \\ \hline
		
		19&1 & 12hrs & 0s & 0s \\ \hline
		
	\end{tabular}	
	\label{table:false_alarm_result}
\end{table}

    


\begin{figure*}[!t]
\centering
\begin{subfigure}{.25\textwidth}
    \includegraphics[scale=0.25]{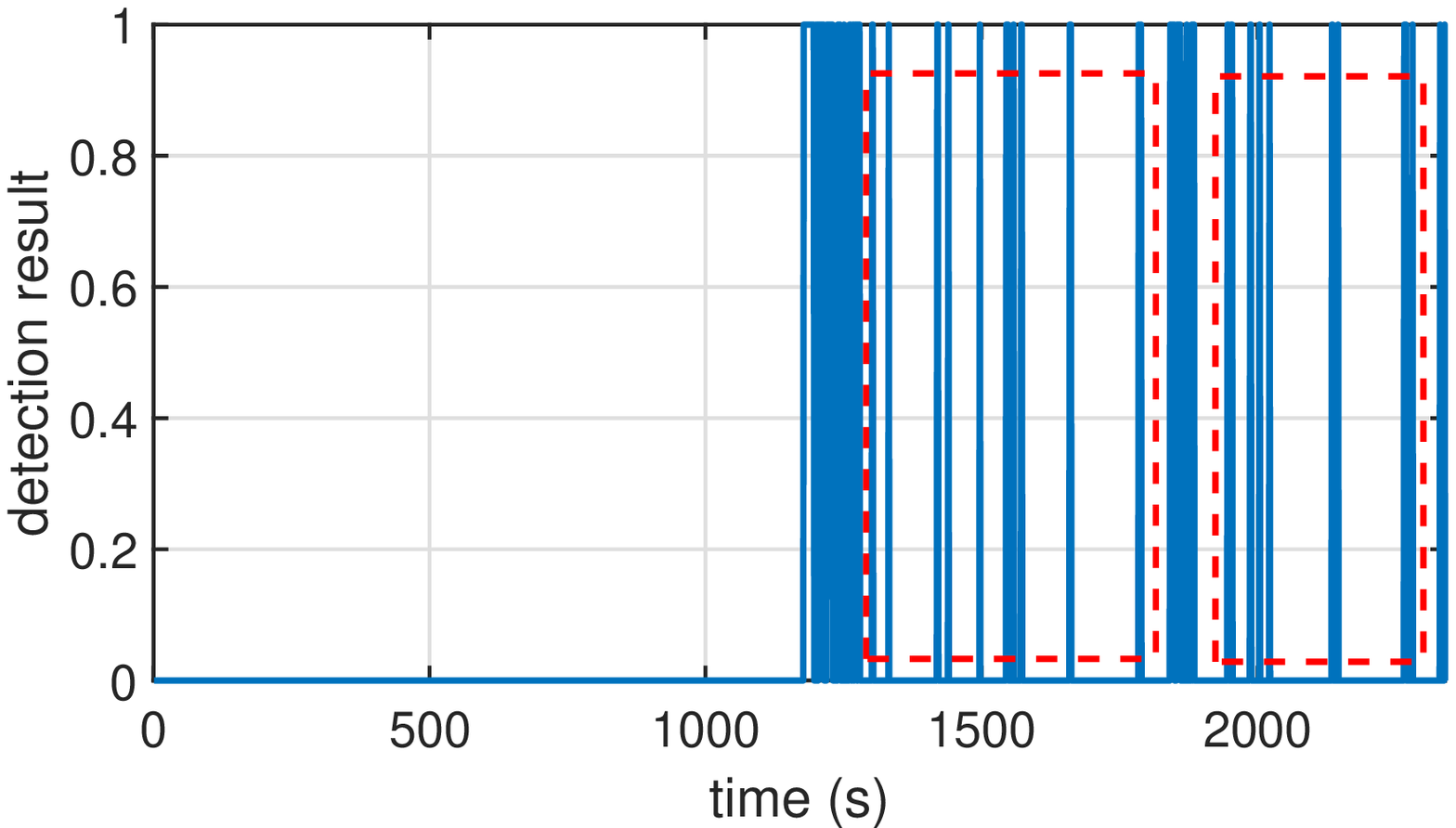}
    \caption{Test 1: CNN}
    \label{fig:test_1_cnn}
\end{subfigure} \hfil
\begin{subfigure}{.25\textwidth}
    \includegraphics[scale=0.25]{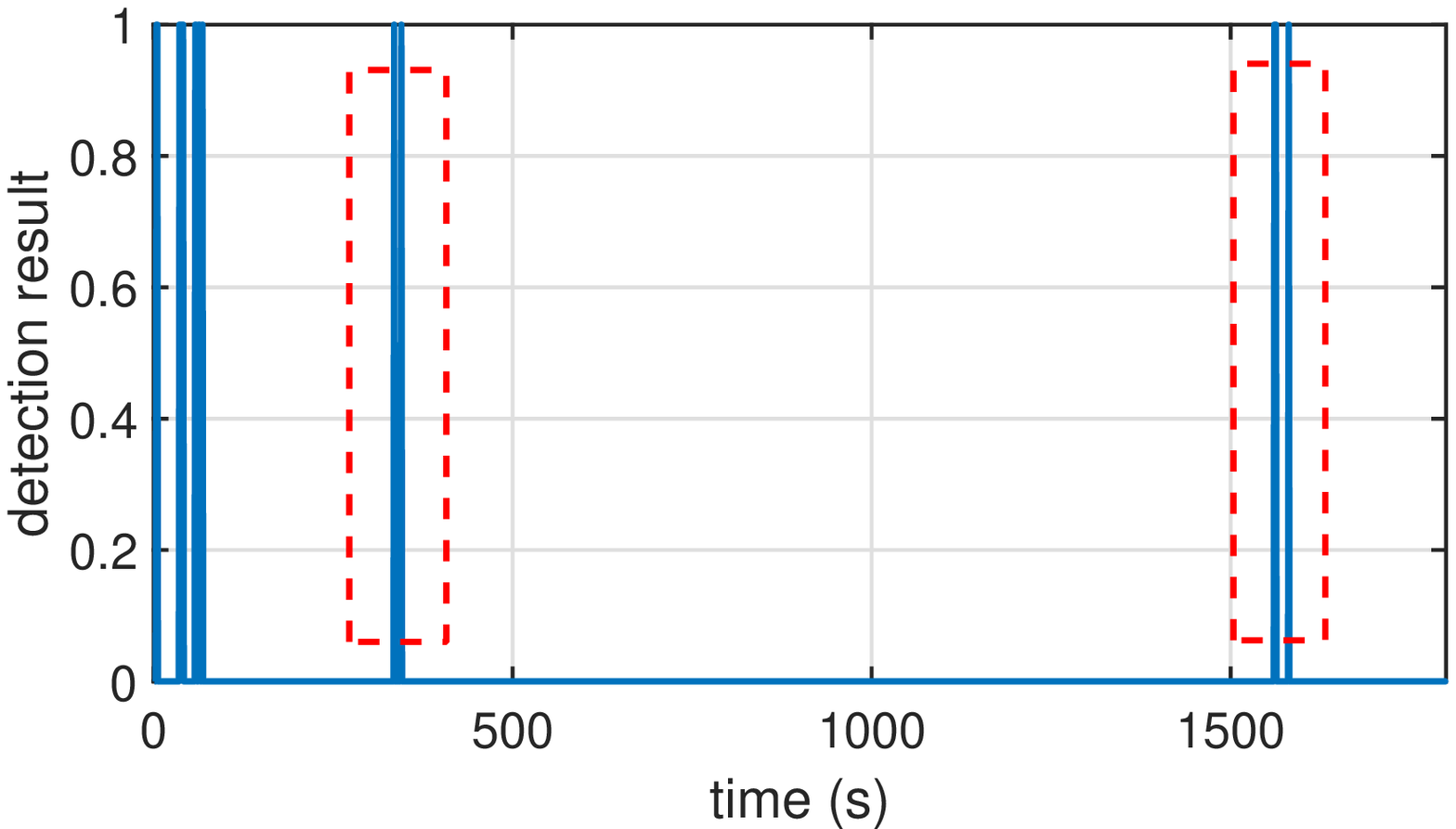}
    \caption{Test 2: CNN}
    \label{fig:test_2_cnn}
\end{subfigure} \hfil
\begin{subfigure}{.25\textwidth}
    \includegraphics[scale=0.25]{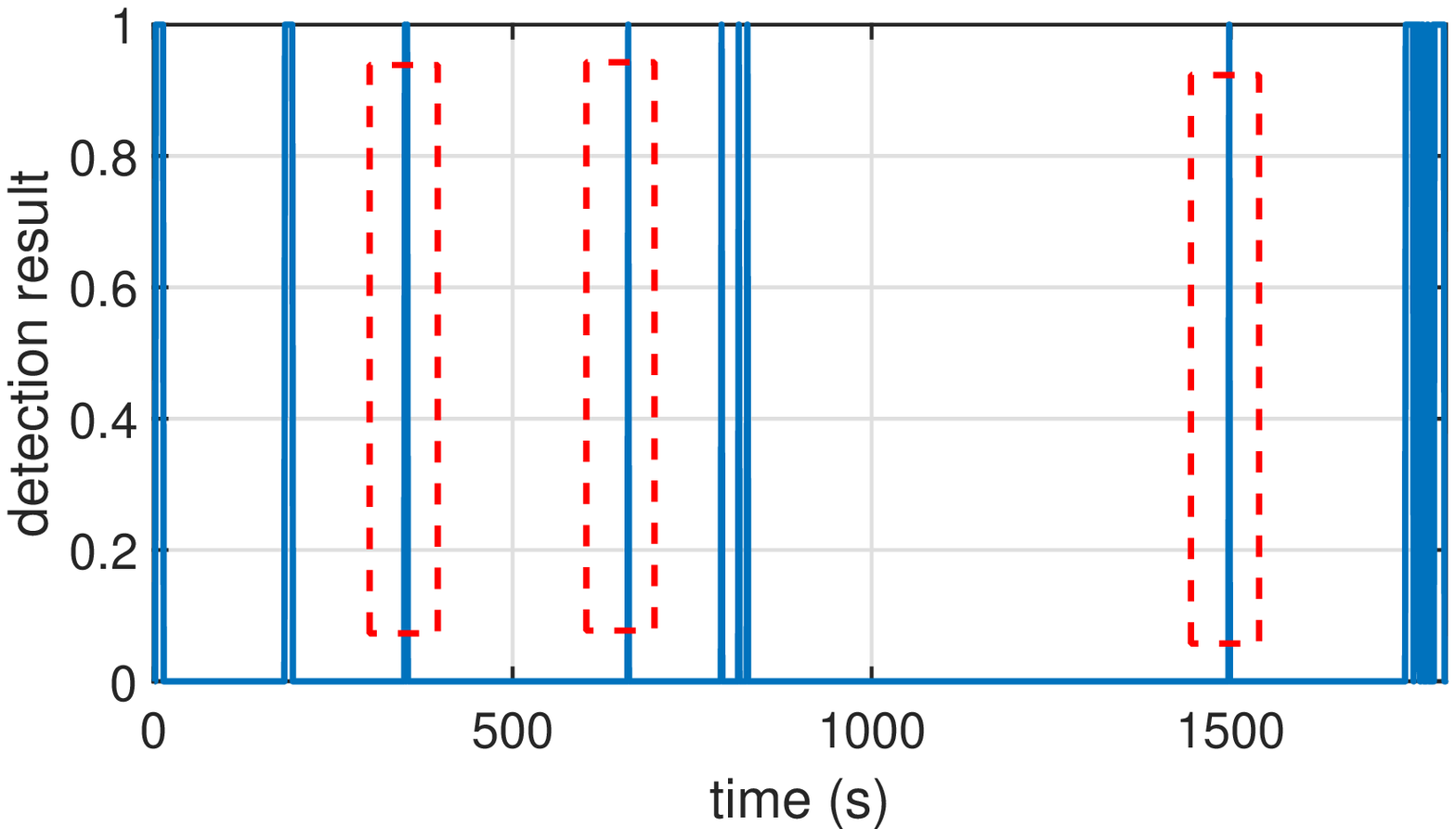}
    \caption{Test 3: CNN}
    \label{fig:test_3_cnn}
\end{subfigure}

\medskip

\begin{subfigure}{.25\textwidth}
    \includegraphics[scale=0.25]{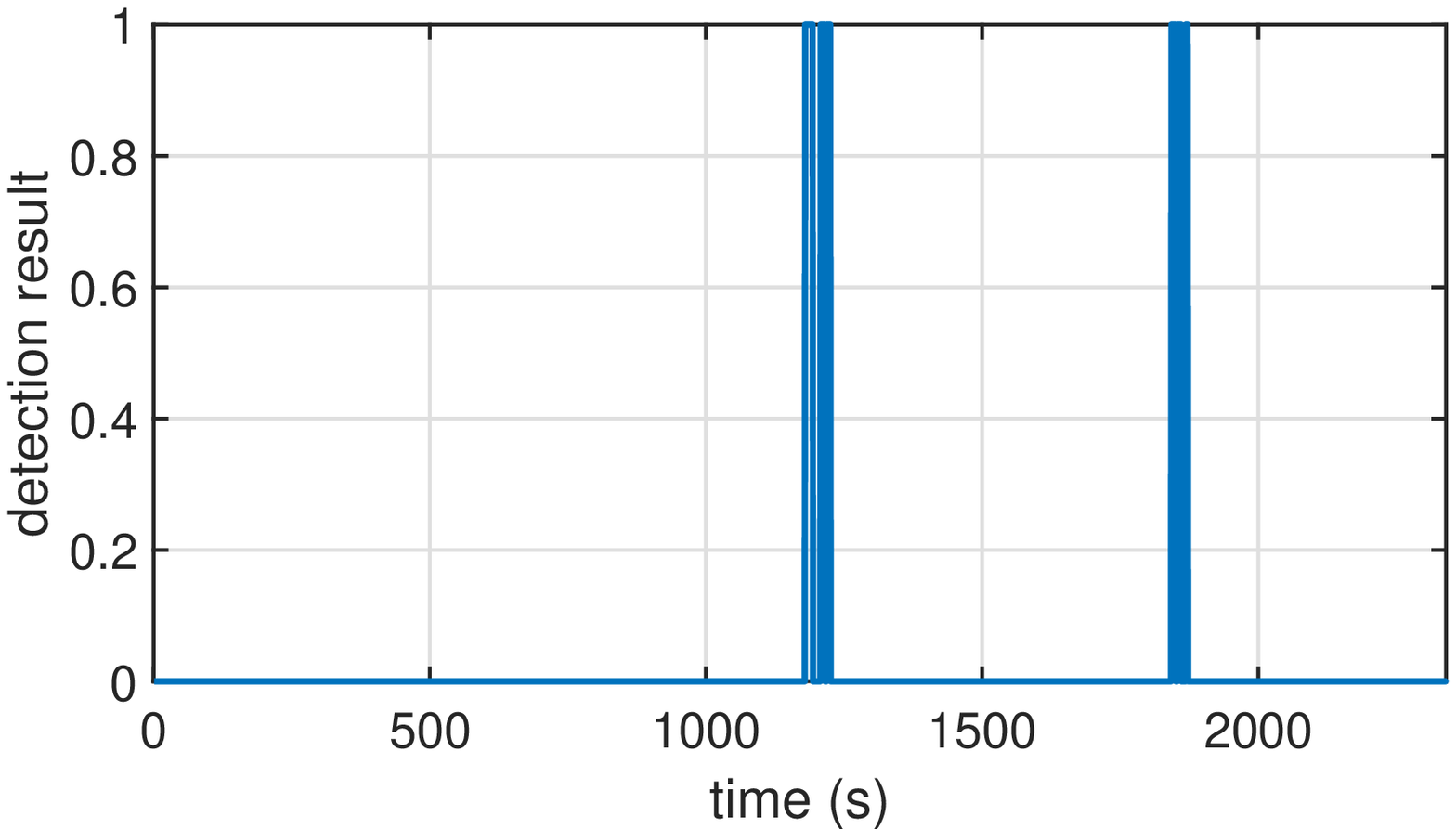}
    \caption{Test 1: PIR}
    \label{fig:test_1_pir}
\end{subfigure} \hfil
\begin{subfigure}{.25\textwidth}
    \includegraphics[scale=0.25]{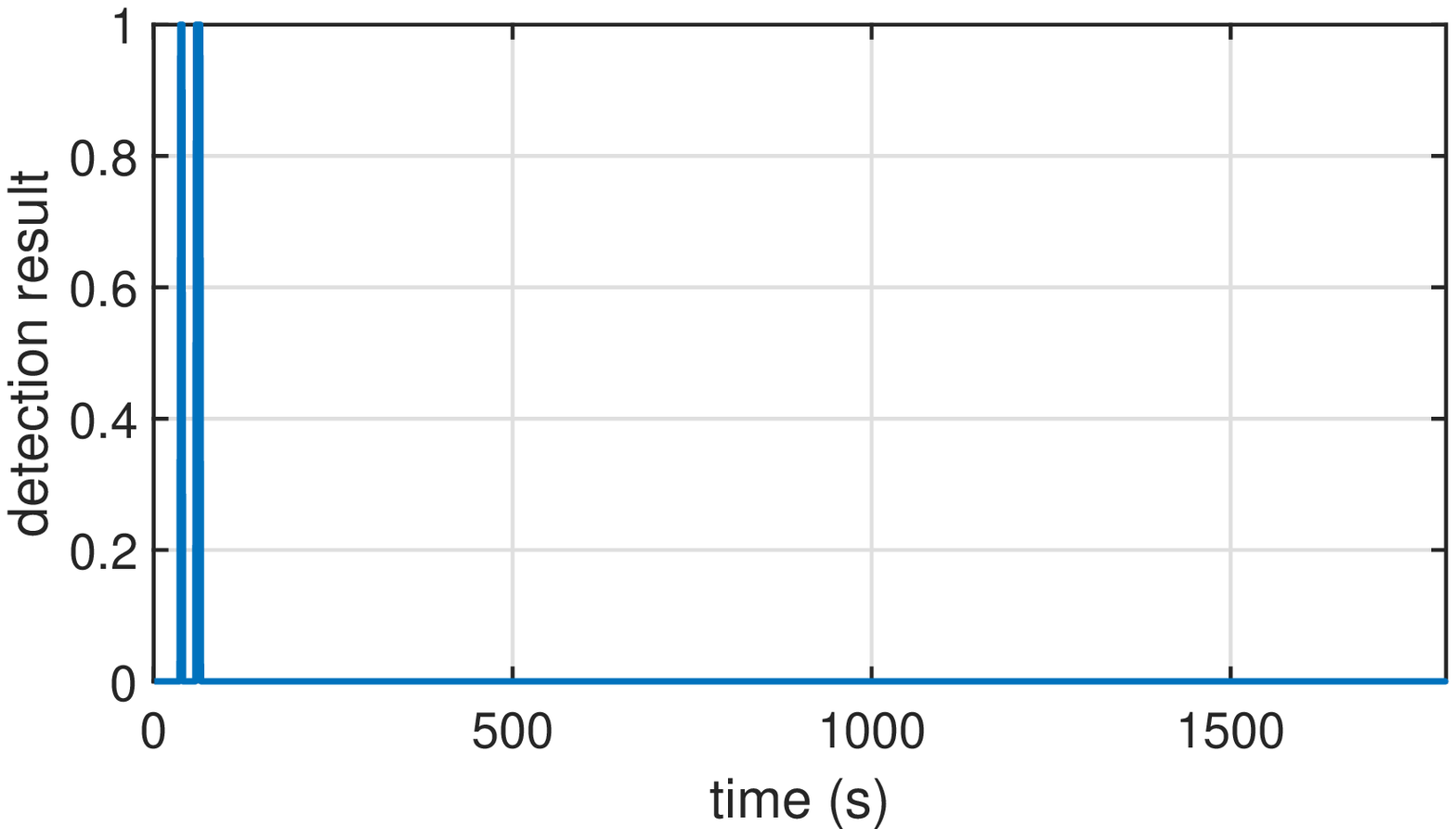}
    \caption{Test 2: PIR}
    \label{fig:test_2_pir}
\end{subfigure} \hfil
\begin{subfigure}{.25\textwidth}
    \includegraphics[scale=0.25]{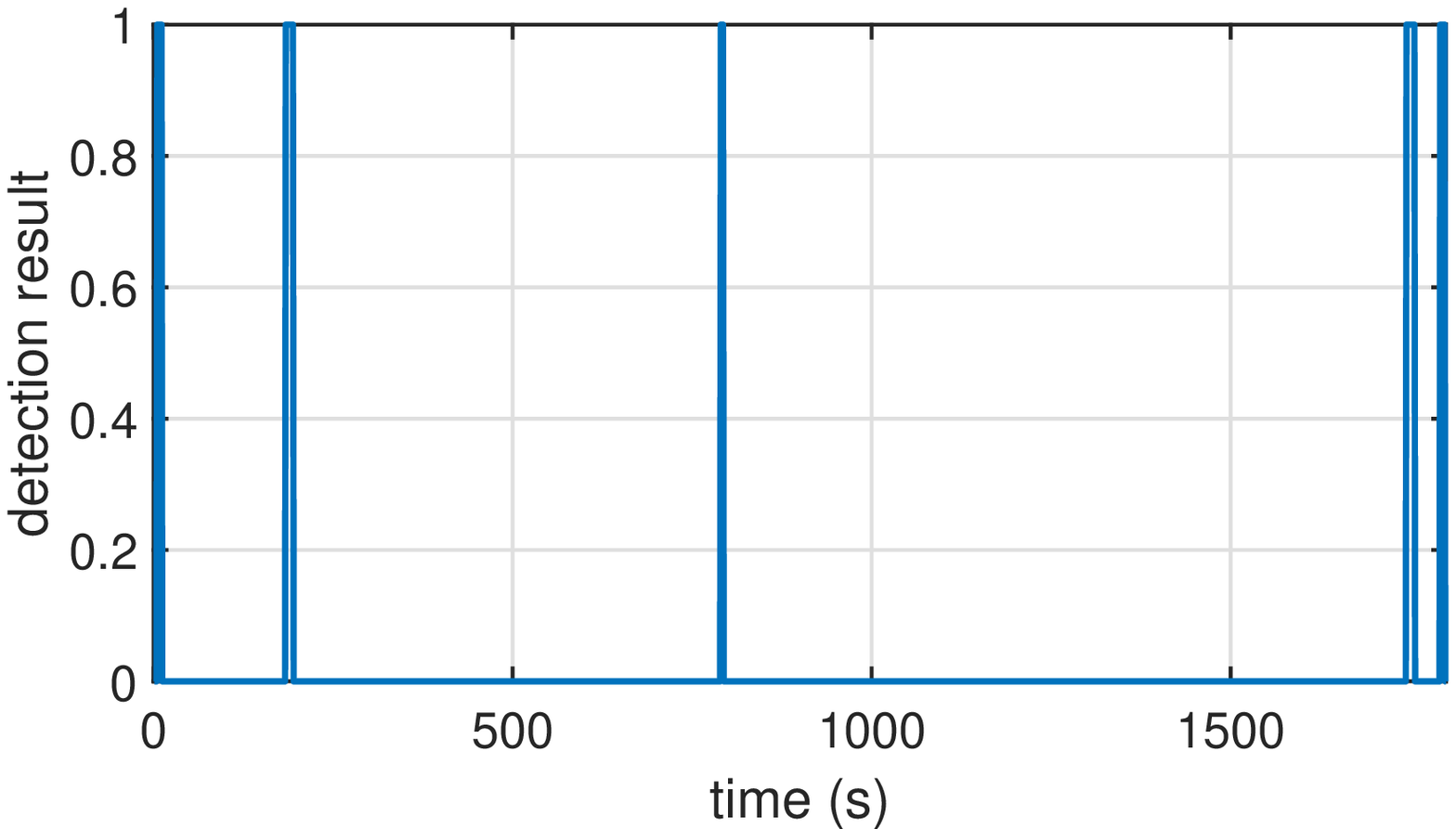}
    \caption{Test 3: PIR}
    \label{fig:test_3_pir}
\end{subfigure}
\caption{Comparison with PIR sensor}
\label{fig:pir_compare}

\end{figure*}




\begin{table}[!t]
	\renewcommand{\arraystretch}{1.5}
	\caption{Presence counts (in seconds)}
	\centering
	\begin{tabular}{| c| c|c|c| c|}
		\hline
		day & test index & duration & CNN & PIR \\ \hline
		\multirow{2}{*}{17} & 1 & 1800s & 212s & 119s \\ \cline{2-5}
		& 2 & 1800s & 76s & 26s \\ \hline
		\multirow{3}{*}{18} & 1& 2340s & 117s & 48s \\ \cline{2-5}
		&2 & 1800s & 19s & 11s \\  \cline{2-5}
		&3 & 1800s & 68s & 41s \\ \hline
		
	\end{tabular}	
	\label{table:sensativity_result}
\end{table}

\subsubsection{Sensitivity test}
This part evaluates the sensitivity of the system to human presence. The experiments are done when people go about with their daily activities in the lab without introducing intentional motions. In most of the time, occupants would just sit in front of their computers and occasionally engaged in normal conversations. Five tests are done in days $17$ and $18$. Duration of each test and presence counts reported by CNN and PIR are summarized in Table~\ref{table:sensativity_result}. Fig.~\ref{fig:pir_compare} shows detection results of $3$ tests done on day $18$. All tests were done with at least one person present in the lab from the beginning to the end except test $1$ on day $18$ as in Figs.~\ref{fig:test_1_cnn} and \ref{fig:test_1_pir} when the lab is empty for the first $20$ minutes. Human activities detected by CNN but not by PIR are marked using red rectangular boxes in Fig.~\ref{fig:pir_compare}. Notice that we only highlight parts when there is not a single positive detection output from PIR sensor for the entire duration of the box. 
For example, at around $1200$s in Figs.~\ref{fig:test_1_cnn} and \ref{fig:test_1_pir}, CNN can detect much longer human presence than the PIR sensor but is not marked in the figure for clarity of presentation. 
To compare the sensitivity of two systems more accurately, we summarize presence counts in Table~\ref{table:sensativity_result}. Each count corresponds to a positive detection for a one second period during the entire test run. WiFi sensing consistently outperforms PIR in all runs.
By cross reference with video recordings, we find that  the presence detected in the highlighted ranges (i.e., those detected by WiFi but not by PIR) in Fig.~\ref{fig:pir_compare} is associated with subtle  human movement, such as stretching while sitting, adjusting sitting postures and conversing with each other without excessive movement. These subtle movements are often missed by PIR but can be easily picked up by WiFi sensing. We emphasize again that model IV is trained with only random walking for label $1$ data, i.e., no small scale motion is included.

It is worth noting that there are still movements missed by both WiFi sensing and PIR. The most important example is when occupants are typing on keyboards but otherwise remain completely still. Such movement appears to be too subtle to be detected by even WiFi sensing. A possible remedy is to deliberately add those keyboard typing measurement data to the motion training set yet it is likely to increase the false alarm rate given the subtlety of such movements.

\subsection{Discussions}
We discuss in this section the impact of various design parameters on the WiFi sensing performance. 

\subsubsection{CSI sampling interval} \label{exp:sampling_interval}
The CSI sampling interval is set at $10$ms in all the experiments reported above. Retraining model I under two more sampling intervals, $20$ms and $40$ms, slight degradation occurs for motion detection but with negligible effect on the presence detection with properly designed post-processing, {\em provided that training and testing are done in the same lab space}. With training and testing done at different labs, slower sampling rate leads to noticeable performance loss. 

\subsubsection{Pre-processing} \label{exp:diss_pre_process}
An important pre-processing step in the proposed system is applying 2-D DFT and 1-D DFT to CSI magnitude images and CSI phase images respectively. We now compare the performance with one that does not use DFT for pre-processing. The CNN architecture for inputs without DFT is modified to achieve best training and test performance to guarantee a fair comparison. In particular, without DFT, the kernel size of the first convolutional layer is changed from $(3,3)$ to $(5,3)$ and the pooling size of the following pooling layers is changed from $(2,1)$ to $(4,1)$ and $(3,1)$ to $(4,1)$ respectively. 
This small adjustment leads to noticeable improvement in false alarm rate while maintaining similar sensitivity in test set compared with the one used for input with DFT. The motivation of increased kernel size to due to the fact that DFT pre-processing localizes CSI dispersion into low frequency region, hence allows the use of significantly smaller kernel size. In the absence of DFT, large kernel size helps capture such channel dispersion due to human motion. 

\begin{figure}[!t]
	\begin{subfigure}[b]{0.5\textwidth}
	\centering
	\includegraphics[scale=0.35]{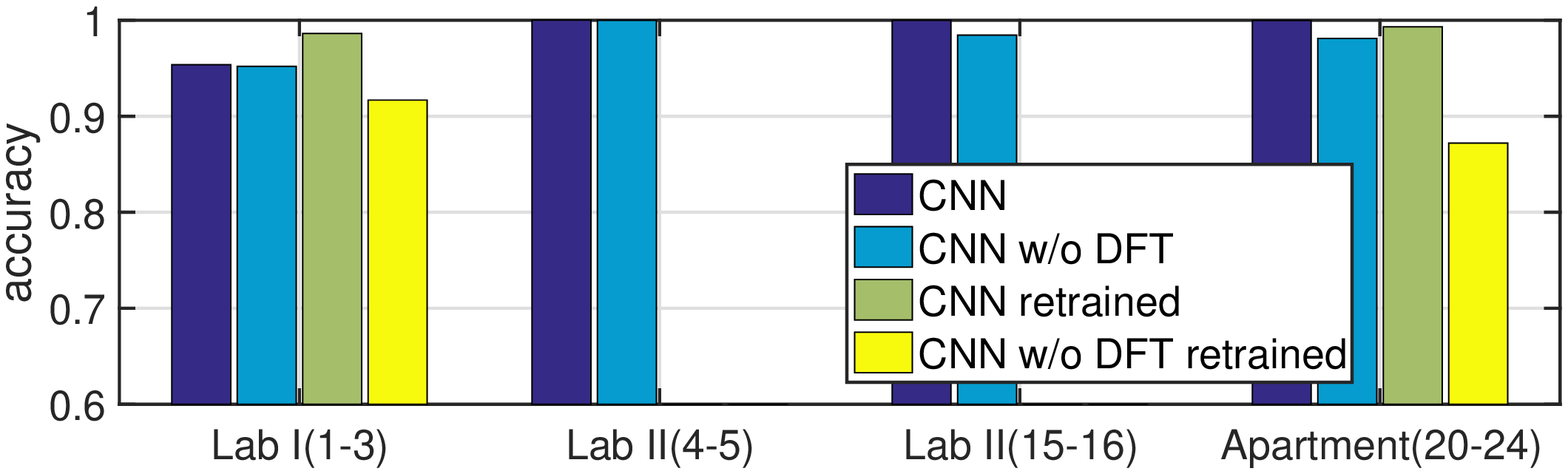}
	\caption{Human free}
	\label{fig:performance_fft_empty}
	\end{subfigure}
	\begin{subfigure}[b]{0.5\textwidth}
	\centering
	\includegraphics[scale=0.35]{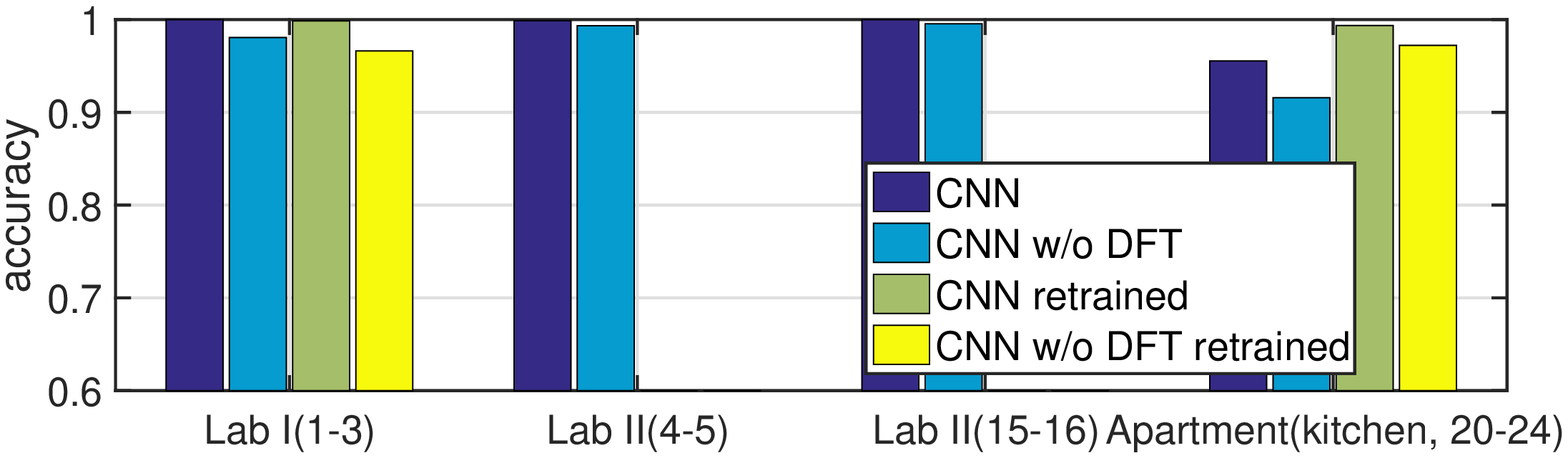}
	\caption{Human motion}
	\label{fig:performance_fft_motion}
\end{subfigure}
\caption{Pre-processing comparison}
\label{fig:performance_preprocessing}
\end{figure}

As in Section~\ref{exp::perf_cmp}, models with and without DFT are evaluated under $4$ test scenarios whose results are given in Fig.~\ref{fig:performance_preprocessing}. Apartment test only reports motions in the kitchen area as performance in other three locations are similar to the lab environment. In each test scenario, the first two bars correspond to models trained with the same training set as model I, while the last two bars for Lab I and apartment show results after retraining as described in Section~\ref{exp::test_different_environment}. In Lab II, both models give comparable results for motion detection, while CNN with DFT has lowest false alarm rate ($1.5\%$ lower false alarm rate in Lab II (day 15-16)). When tested in Lab I, both models have noticeable degradation in false alarm performance. With retraining using human free data from day $3$, the CNN with DFT has significantly lower false alarm rate while the CNN without DFT does not have any improvement. For the kitchen area in the apartment, both models give higher false negatives than the lab environment. The simple retraining method does not help CNN without DFT since it increases the motion sensitivity in the kitchen at the expense of further elevating false positive in a human free environment (false alarm rate rises by $10.9\%$). With DFT, however, learning becomes more immune to CSI features \textit{not} attributed to human motions, leading to noticeable improvement in performance when retrained with additional data. 

\subsubsection{Input} \label{exp:input_cmp}
The proposed system harvests presence information from  CSI magnitude and phase in an explicit manner. In this part, we investigate the impact of different inputs on detection performance: a) CSI magnitude only; b) CSI phase only; c) complex CSI input without explicit decomposing into magnitude and phase. For c), we combine the last two spatial dimensions of the 4-D CSI array $\mathbf{X}\in\mathbb{C}^{128\times14\times3\times3}$ into one and apply normalization as in \eqref{eq:time_div} but to complex entries. The resulting real and imaginary components are then stacked up along the last dimension, resulting in inputs to a CNN model of shape $128 \times 14 \times 18$. The sizes of convolutional/pooling layers are adjusted as in Section~\ref{exp:diss_pre_process} for a fair comparison due to enlarged input size.

\begin{figure}[!t]
\centering
	\begin{subfigure}[b]{0.5\textwidth}
	\includegraphics[scale=0.35]{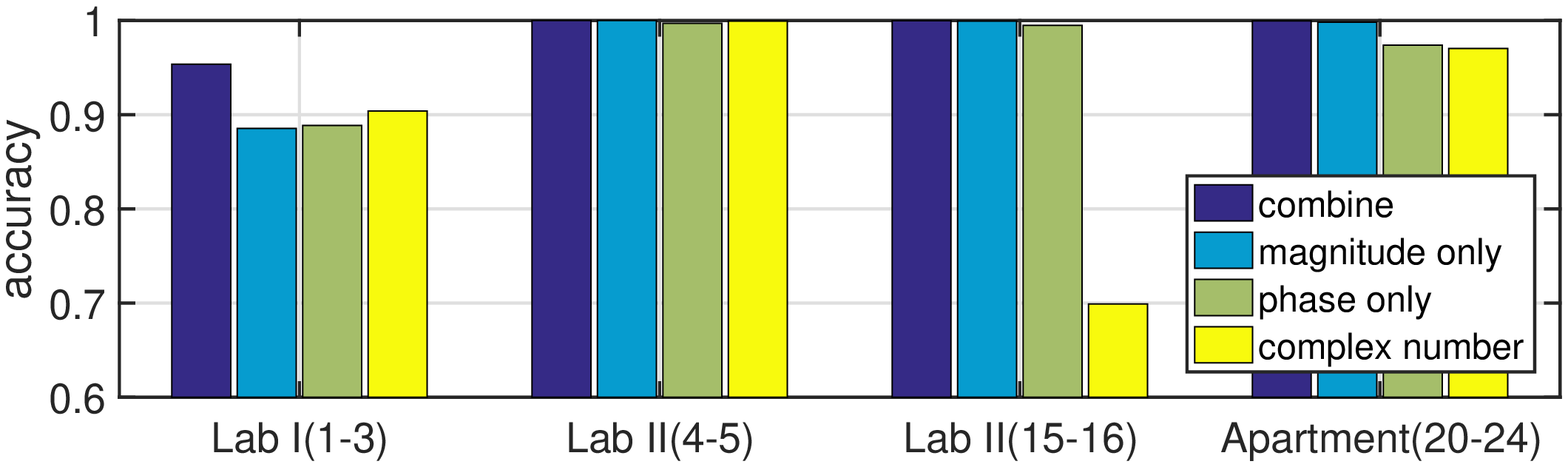}
	\caption{Human free}
	\label{fig:performance_input_empty}
	\end{subfigure}
	\begin{subfigure}[b]{0.5\textwidth}
	\includegraphics[scale=0.35]{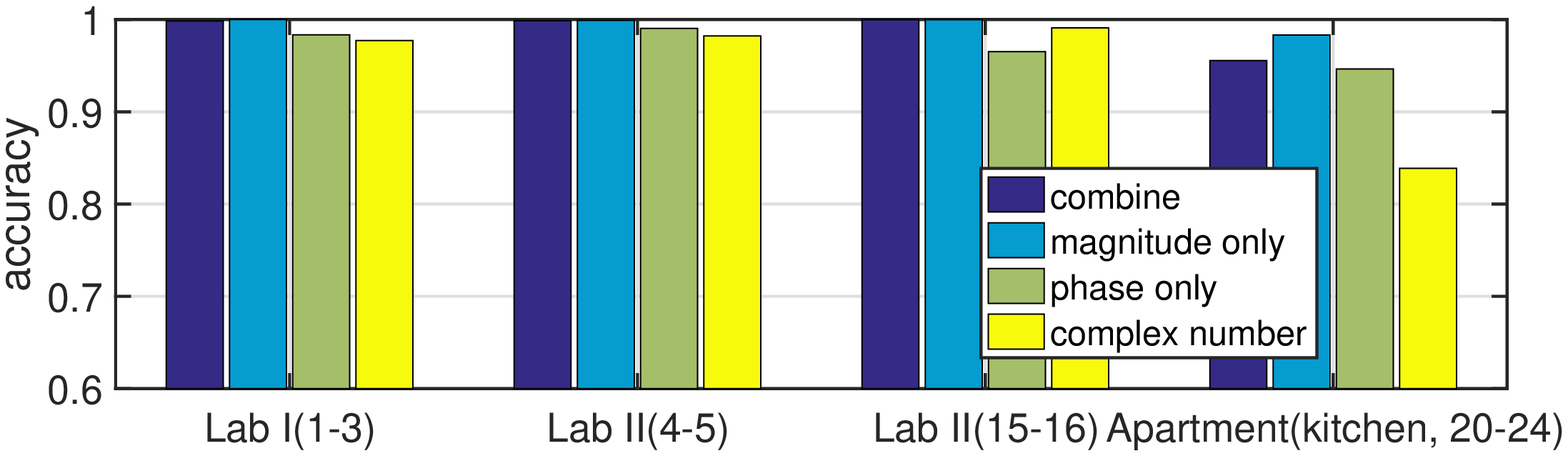}
	\caption{Human motion}
	\label{fig:performance_input_motion}
\end{subfigure}
\caption{Input comparison}
\label{fig:performance_input}
\end{figure}

The results of the three models plus Model I are given in Fig.~\ref{fig:performance_input}. With information from Lab II only, as seen in Fig.~\ref{fig:performance_input_empty}, Model I, which explicitly uses both CSI magnitude and phase, results in the most robust model in a human free environment. In contrast, the performance using complex CSI input without decomposing into magnitude and phase varies wildly for different test days. While it has low false positive rate on days 4 and 5, it has highly elevated false positive rate on days 15 and 16. When tested in Lab I, which is different from where the training data were collected, the three models all lead to higher false positive detection compared with Model I.

Motion detection results are given in Fig.~\ref{fig:performance_input_motion} where, as before, for the apartment scenario, only the test result in the kitchen is plotted. 
Complex CSI input is consistently worse than Model I and CSI magnitude only in all four test scenarios. 
Overall, CSI phase only is not as sensitive to motions as CSI magnitude only. Indeed, the performance of Model I in the apartment is negatively affected by the CSI phase, which can be compensated by retraining Model I with data collected for kitchen from day $20,22$ (c.f. model III in Table~\ref{table:training_data_composition}). These additional training leads to an improvement in detection performance 
from $95.55\%$ to $99.36\%$ with combined magnitude and phase input.

\subsubsection{Architecture design}\label{exp:architecture_design}

\begin{figure}[!t]
	\begin{subfigure}[b]{0.5\textwidth}
	\centering
	\includegraphics[scale=0.35]{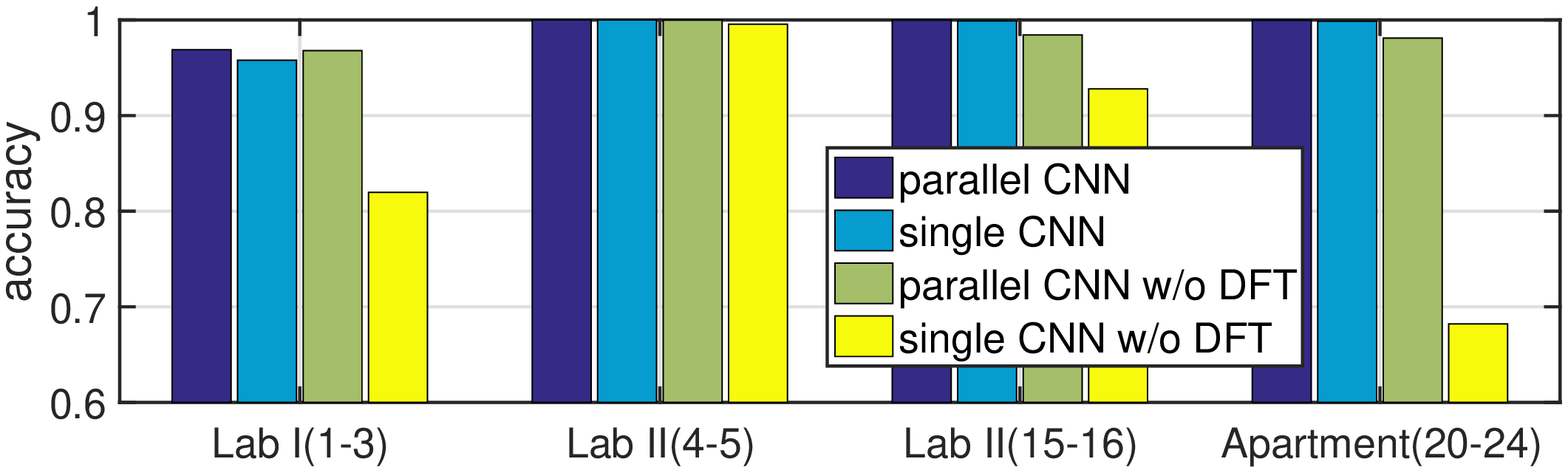}
	\caption{Human free}
	\label{fig:performance_structure_empty}
	\end{subfigure}
	\begin{subfigure}[b]{0.5\textwidth}
	\centering
	\includegraphics[scale=0.35]{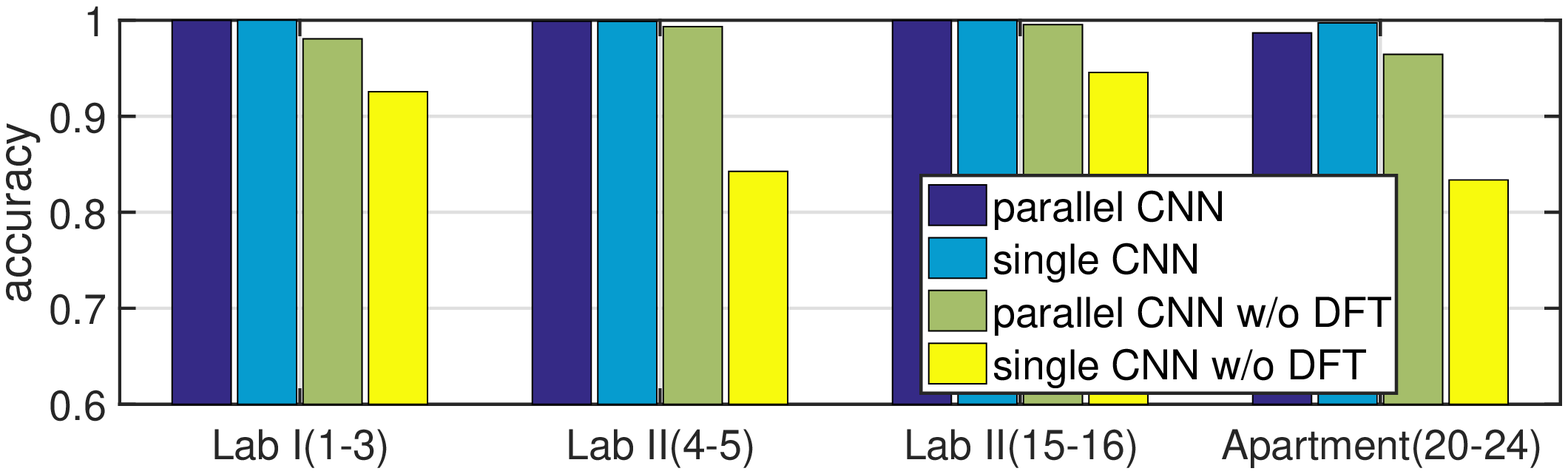}
	\caption{Human motion}
	\label{fig:performance_structure_motion}
\end{subfigure}
\caption{Architecture comparison}
\label{fig:performance_architecture}
\end{figure}

The proposed CNN uses a parallel structure to extract information from CSI magnitude and phase images separately. An alternative architecture is to use a single conventional CNN by concatenating the magnitude and phase images along the last axis. Thus, the inputs to the CNN are of size $50\times 14\times 15$. The performance of the parallel CNN and the single CNN are given in Fig.~\ref{fig:performance_architecture}. Notice that in this part, detection accuracy in the apartment is averaged over all four locations since performance of the models differ in most areas. With training data from Lab II only, these two models have comparable performance in Lab II. When tested in Lab I, the parallel CNN has slightly less false positives in the human absent environment; in the apartment, however, the single CNN results is more sensitive to human motions. Overall, the parallel CNN and the single CNN have comparable performance. 

Finally, we evaluate the impact of pre-processing for the two different CNN architectures. Specifically, we evaluate the parallel and single CNN architectures with input without DFT on the CSI images. The sizes of convolutional/pooling layers are adjusted accordingly as in Section~\ref{exp:diss_pre_process} to ensure a fair comparison. As seen in Fig.~\ref{fig:performance_architecture}, there exists a noticeable performance gap between the parallel and single CNN without DFT for all test scenarios. Without DFT, the single CNN architecture tends to overfit training data, thus yielding lower accuracy in test data.  

%% file: modules/conclusion.tex

In this paper, a passive WiFi sensing system is proposed for indoor occupancy detection. The system exploits motion induced variation in both magnitude and phase of the CSI. 
A parallel convolutional neural network architecture is adopted to harvest occupancy information in CSI estimates along temporal, frequency, and spatial dimensions. With judicious pre-processing to remove hardware/system impairments and post-processing to infer presence information from motion detection output, the proposed learning system provides a viable and promising alternative for real time occupancy detection. Extensive experiments were conducted using commercial off-the-shelf WiFi devices. It was demonstrated that system is much more sensitive to human presence than PIR sensors and exhibit desired robustness against environment variation compared with existing RF based presence detection systems.  

A key challenge for presence detection is the calibration of human motion to achieve balance between sensitivity to human presence and false alarm. The collection and use of data with human motion has an outsized influence on the presence detection performance. Future work will explore presence detection using only training data corresponding to empty rooms for the desired robustness. Learning approaches such as universal hypothesis test and one-class SVM may prove useful alternatives than deep learning.